\ifdefined\XeTeXrevision\else\pdfoutput=1\fi
\documentclass{lynnreal}
\usepackage{amsmath,amssymb}
\usepackage{booktabs}
\usepackage{array}
\titleformat*{\subsubsection}{\normalsize\bfseries\sffamily\color{lynnfg}}
\microtypesetup{expansion=false}
\newcommand{\method}{LynnReal-Omni}

\title{{\LARGE LynnReal-Omni: Native multi-modal Video Generation for Agentic Visual Workflows}}
\author{\centering \textbf{LynnReal AI*}}

\contribution[*]{Lynnreal-Omni contributors are listed at the end of the report.}

\abstract{
Video diffusion models are stochastic and hard to control: precise content often requires repeated sampling without guaranteed success, and long-horizon scenes drift in appearance, interactions, and temporal coherence. Agentic visual creation provides explicit references, editable 3D scenes, or executable game states for stable control, but does not by itself guarantee high object or character fidelity. Combining the two can enable stable, high-quality generation. To realize this combination, we present \textbf{LynnReal-Omni}, a native multimodal video generation framework built on a 32B shared multimodal diffusion transformer that unifies text-to-video, image-conditioned generation, reference-guided generation, structural control, editing, degraded video restoration, and long-video generation. It accepts heterogeneous visual inputs—appearance references, editable 3D renders, and game recordings—allowing agents to compose visual conditions within a unified model. We also train a dedicated 27B \textbf{Flash} shared multimodal diffusion transformer for \textbf{real-time rendering}. We build a systematic data pipeline for video cleaning, subject association, multimodal annotation, and aligned control construction, yielding a curated corpus of multi-shot audiovisual segments, and introduce \textbf{MSAVP}, a 100-prompt, 20-metric evaluation design that separates instruction following, generating plausibility, visual quality, temporal behavior, and audio coordination. LynnReal-Omni-Flash further reduces inference cost through model and decoding acceleration, including a \textbf{lightweight VAE decoder}; on one H100, warm generation and decoding of a 22-frame 540p video take  \textbf{843 ms} with LynnReal-Omni and  \textbf{377 ms} with Flash. These results provide a foundation for real-time streaming video generation, making LynnReal-Omni a unified, controllable, and efficient basis for agentic visual creation.
}
\date{September 2026}
\lynndata[Version]{Technical report; September 2026}
\lynndata[Code]{\url{https://github.com/LynnReal-AI/LynnReal-Omni}}
\lynndata[Demo]{\url{https://www.youtube.com/watch?v=P5Bl2mriEmk}}
\lynndata[Flash]{\url{https://huggingface.co/stdstu123/LynnReal-Onmi-flash-beta-0.1}}
\lynndata[Standard]{\url{https://huggingface.co/stdstu123/LynnReal-Onmi-beta-0.1}}
\lynndata[Light-vae]{\url{https://huggingface.co/stdstu123/LynnReal-Onmi-light-vae}}
\begin{document}
\maketitle
\tableofcontents
\begin{figure*}[!h]
\centering
\includegraphics[width=\textwidth]{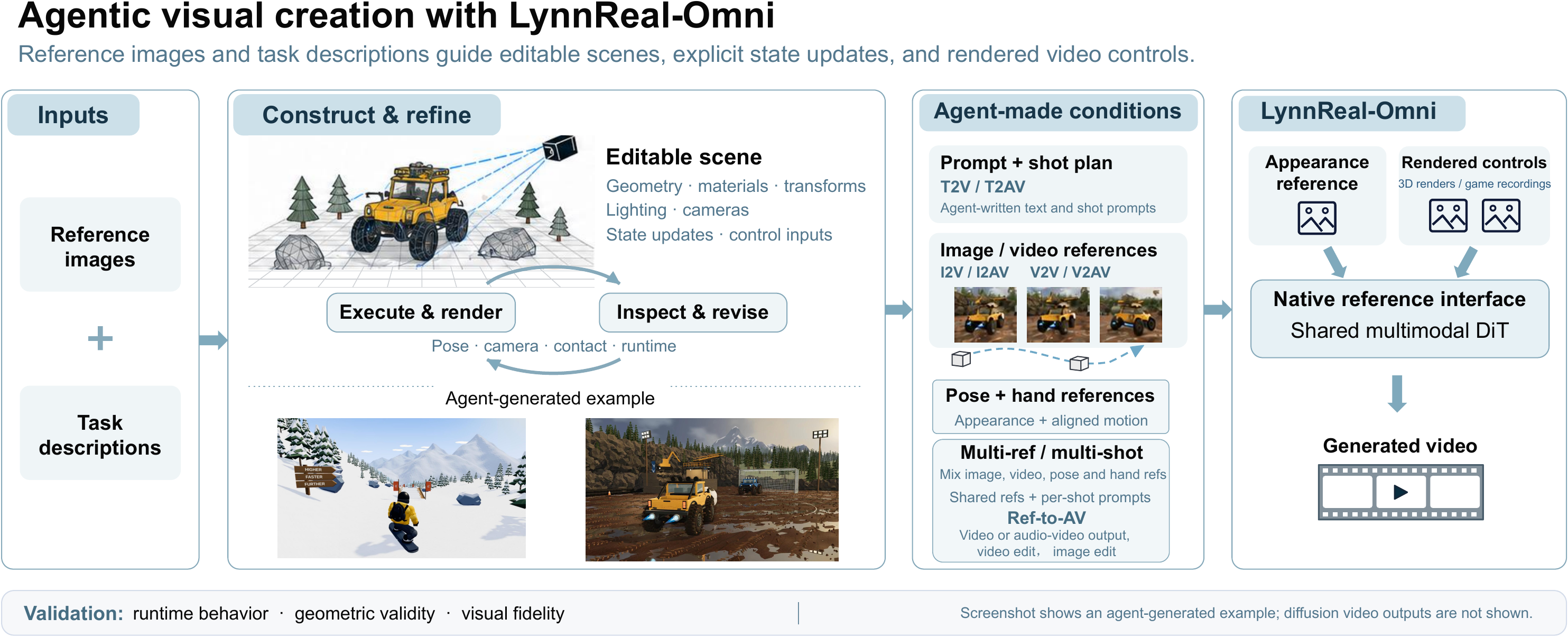}
\caption{Agent workflows. Image-based scene reconstruction and agent-written low-poly games produce distinct visual-control streams. Prompt refinement, image generation, and first-frame editing provide appearance controls. The video model receives these controls through its native reference interface.}
\label{fig:agent}
\end{figure*}

\section{Introduction}
Video diffusion models~\citep{ho2022video,blattmann2023align,polyak2024movie,wan2025,valevski2024diffusion,chen2025sana,chen2024videocrafter2,ceylan2023pix2video,harvey2022flexible,wang2024motionctrl,zhang2024cameractrlii,he2024cameractrl} have achieved remarkable visual fidelity, but they remain difficult to control. Generation is stochastic~\citep{ho2020ddpm}, precise content often requires repeated sampling without any guarantee of success, and long-horizon scenes tend to drift in appearance~\citep{lu2026reward,cui2025self,huang2025self}, object interactions, and temporal coherence. These limitations make it hard to use video diffusion models as reliable rendering and generation engines in workflows that demand explicit and repeatable control. Agentic visual creation~\citep{chen2026code,ye2026genclaw,openai2026gpt6astra} offers a complementary source of control: an agent can produce reference images, construct editable 3D scenes with camera trajectories, or write executable games with controllable objects and collision rules, thereby making scene geometry and motion explicit. Such conditions stabilize the generation process and reduce the need for repeated sampling. However, agentic control alone does not guarantee high-fidelity object or character appearance. Combining agentic visual creation with video diffusion therefore provides a promising path toward stable, high-quality generation.

Realizing this combination, however, requires general-purpose video models that can understand and combine diverse multimodal references and conditions while maintaining coherent appearance, motion, interactions, audio, and long-term consistency. Existing approaches fall short in several important respects: 

\noindent (1) \textbf{Fragmented task-specific pipelines.} Task-specific pipelines for text-to-video, video-to-video, image-conditioned generation, reference-guided generation, structural control, editing, and long-video generation have advanced largely in isolation, so agents must compose multiple models and ad hoc interfaces, which increases engineering cost and often produces inconsistent behavior across tasks; 

\noindent(2) \textbf{Long-horizon inconsistency.} Long-video methods often suffer from appearance drift~\citep{lu2026reward,cui2025self,huang2025self}, identity changes, and photometric artifacts; 

\noindent(3) \textbf{Incomplete evaluation protocols.} Existing evaluation protocols such as VBench~\citep{huang2024vbench} and VideoPhy~\citep{bansal2024videophy} separate some dimensions of video quality, but they do not comprehensively cover multi-shot continuity, action binding, physical plausibility, controllability, and audio quality in a single protocol, and metrics are often aggregated across incompatible scales, hiding failures in specific dimensions;

\noindent(4) \textbf{Decoding bottleneck.} Video diffusion models are often assumed to be dominated by the denoising transformer, but few-step distillation sharply reduces denoising cost and shifts the bottleneck to the VAE decoder, which becomes the dominant inference cost. Decoder acceleration is therefore essential, yet it must be evaluated jointly with denoising, since reducing decoder compute can affect temporal phase or reconstruction quality. 

To address these limitations, we present \textbf{LynnReal-Omni}, a native multimodal video generation framework built on a shared multimodal diffusion transformer. Its design addresses the above gaps in several ways:
\begin{itemize}
    \item Rather than treating each task as a separate pipeline, LynnReal-Omni unifies text-to-video, image-conditioned generation, reference-guided generation, structural control, editing, and long-video generation within a single model, replacing fragmented task-specific stacks with a shared denoiser and task-specific input layouts.
    \item A native task representation explicitly encodes modality identity, temporal coordinates, noise levels, and output targets, allowing appearance references, frame-aligned controls, editable 3D renders, game recordings, and causal history to retain their distinct roles within a shared multimodal transformer.
    \item A systematic data pipeline for video cleaning, subject association, multimodal annotation, and aligned control construction yields a curated corpus of multi-shot audiovisual segments, providing source-linked training units with verified conditioning assets.
    \item We introduce \textbf{MSAVP}, a 100-prompt, 25-metric benchmark for evaluating semantic alignment, visual quality, temporal consistency, physical plausibility, controllability, and audio quality, which separates semantic compliance from observed physics and retains distinct visual, temporal, and audio measures.
    \item LynnReal-Omni is optimized for practical deployment, and \textbf{LynnReal-Omni-Flash} reduces inference cost through model and decoding acceleration, including a lightweight VAE decoder, with warm generation and decoding of a 22-frame 540p video taking 909\,ms on one H100 for LynnReal-Omni and 591\,ms for Flash.
\end{itemize}

\section{Related Work}
\subsection{Native multi-modal generation.}
Video foundation models such as CogVideoX, HunyuanVideo, and Wan combine spatiotemporal latent compression with scalable diffusion transformers~\citep{yang2024cogvideox,kong2024hunyuanvideo,wan2025}. LTX-2~\citep{hacohen2026ltx} extend generation to synchronized audio and video through cross-modal interaction, while VACE unifies reference-conditioned generation and video editing through a shared conditioning interface~\citep{jiang2025vace}. Our implementation builds directly on MiniMax-H3~\citep{minimax2026h3}, which provides a joint video--audio transformer, modality-specific codecs, and native keyframe and reference interfaces. Building on this backbone, we integrate image references, motion controls, editing, and long video generation while preserving task-specific token ordering, modality labels, and temporal positions. We distinguish video-only continuation from joint audiovisual generation and evaluate these interfaces together with few-step inference.

\subsection{Distribution matching distillation.}
Distribution Matching Distillation (DMD) trains one-step generators by matching teacher and student output distributions~\citep{yin2024dmd}. DMD2 removes the paired regression requirement and improves training through two-time-scale updates, adversarial supervision, and inference-matched multi-step training~\citep{yin2024dmd2}. Subsequent work extends distribution matching along complementary directions: TDM aligns intermediate trajectory distributions for flexible few-step sampling~\citep{luo2025tdm}; Self Forcing trains on autoregressive student rollouts to reduce exposure bias~\citep{huang2025self}; and Reward Forcing introduces rewarded distribution matching to improve motion dynamics~\citep{lu2025rewardforcing}. More recently, Salt combines self-consistent denoising updates with cache-aware training to improve low-step video generation~\citep{ge2026salt}. Our work applies few-step distillation to both standard and Flash variants, emphasizing multimodal control preservation and measured end-to-end efficiency under their respective deployment configurations.

Trajectory distribution matching~\citep{luo2025tdm} further motivates supervising short student transitions against the teacher distribution. Our standard and Flash paths have different deployment topologies and are evaluated with their respective trained configurations. Depth reduction, token reduction, quantization, operator fusion, and decoder replacement change different parts of the cost. Their effects require separate ablations: a reduction in parameter count does not by itself predict whole-pipeline latency, and a numerically exact kernel improvement differs from a quality-sensitive approximation.

\subsection{Physical and audiovisual evaluation.}
VBench separates several dimensions of video quality~\citep{huang2024vbench}. VideoPhy explicitly tests caption adherence and physical commonsense~\citep{bansal2024videophy}. Our MSAVP protocol similarly keeps semantic, physical, visual, and temporal judgments separate and adds structured accounting for multishot action, binding, sound semantics, and event timing. Specialist measurements provide evidence for cut locations and appearance, including TransNetV2~\citep{soucek2020transnetv2} and MUSIQ~\citep{ke2021musiq}; they cannot substitute for observing an interaction. We retain metric-level applicability counts and original scoring units alongside a hierarchical six-family aggregate, so the overall score does not replace the underlying evidence.

\subsection{Agentic visual creation.}
  Early LLM-based agents orchestrated image generation and editing
  through prompts and tool calls~\citep{wu2023visualchatgpt}.
  Multimodal backbones such as GPT-4o and Gemini 1.5 subsequently
  enabled visual inspection and reasoning over reference images
  and extended video contexts, supporting more elaborate image
  editing and video planning
  workflows~\citep{openai2024gpt4o,reid2024gemini15}.
  Claude 4 further strengthened sustained coding and tool use,
  extending agentic creation toward executable games and interactive
  applications~\citep{anthropic2025claude4}.
  More recently, GPT-6 Astra supports complex workflows combining
  reasoning, coding, and computer use~\citep{openai2026gpt6astra}.
  However, producing detailed animated scenes still requires
  substantial downstream work in geometry, materials, rigging,
  animation, and rendering.
  Stronger agent backbones do not eliminate these production costs,
  and the cited advances do not establish real-time,
  end-to-end creation of finely modeled animated content.
  This motivates a complementary workflow in which agents construct
  lightweight scenes or playable game prototypes, while a fast
  video diffusion model supplies detailed visual appearance.
  
\section{Data Processing}
\label{sec:data}
\subsection{Overview: from public videos to high-quality multishot data}

\begin{figure}[h!]
\centering
\includegraphics[width=\textwidth]{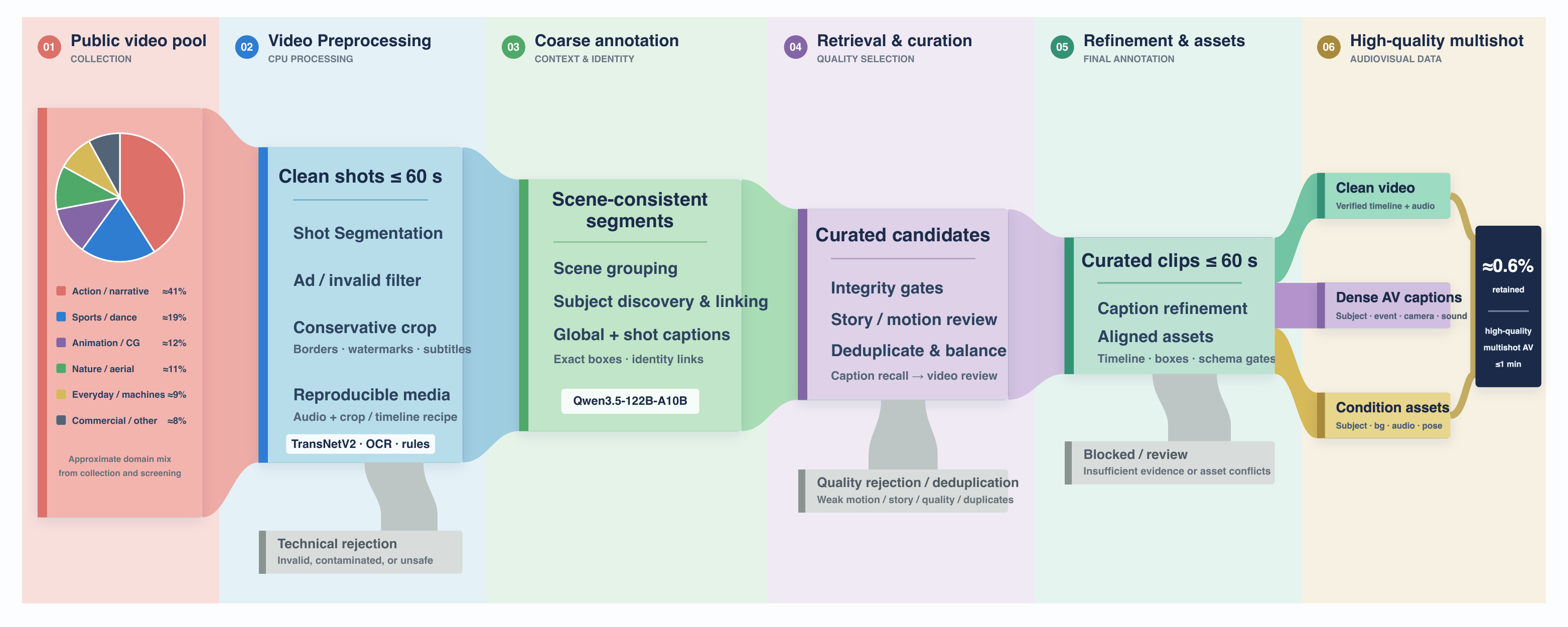}
\caption{The data processing and annotation pipeline. The pie chart shows an approximate content distribution obtained from collection categories and quality-screening labels; percentages are rounded. }
\label{fig:data-pipeline}
\end{figure}

Our goal is to turn diverse public videos into clean and well-described multishot audiovisual clips. The collected videos cover story and action, sports and performance, animation and computer graphics, nature and aerial views, everyday activities and machines, and commercial or other content. We estimate the mixture shown in Figure~\ref{fig:data-pipeline} by combining the categories used during collection with the content types found during quality screening. After all processing stages, approximately 0.6\% of the original storage footprint remains as high-quality data, and every retained clip is no longer than one minute.

\paragraph{Pipeline structure.} The pipeline has five main components. We first detect shot boundaries, segment the videos into shots, and remove visible contamination; then group shots by scene and identify recurring subjects; describe each same-scene multishot clip of at most one minute with both a clip-level caption and detailed per-shot captions, and select clips with clear motion, coherent events, and good visual quality; prepare subject, background, pose, and audio information; and finally produce detailed captions with automatic consistency checks.

\subsection{Video Preprocessing}

\paragraph{Shot Segmentation.} This stage converts raw videos into physically reliable shots. TransNetV2~\citep{soucek2020transnetv2} proposes possible boundaries with a low threshold of 0.1, which keeps recall high at this stage. We then compare the frames on both sides of each candidate using color distribution, brightness, white-pixel ratio, and edge structure. A candidate is rejected when the change can be explained by a flash, camera shake, a fast pan, or a moving object that briefly covers the frame. A cut is retained when the visual evidence supports a real change of shot, including a cut made during a continuous action. Short shots are kept when the evidence supports them, while long videos without cuts are divided into continuous windows of at most 60 seconds without dropping frames or audio.

\paragraph{Text and border cleaning.} PP-OCR~\citep{du2020ppocr} detects and recognizes text in sampled frames. The text is used both to identify text-heavy advertisements and to propose watermark and subtitle regions. Black borders are estimated from brightness and variation across several frames, which prevents a dark scene from being mistaken for a border. We choose the smallest crop that removes the detected regions and reject any crop that would remove more than 25\% of the image. If a safe crop does not exist, the clip is kept for later review or rejection instead of being altered by unconstrained image generation. Static watermarks are identified by repeated text at a stable edge position across the full video, whereas subtitles are identified by repeated text within a shot, usually near the top or bottom. After cropping, we finally retain cleaned shots with no visible contamination.

\subsection{Coarse annotation}

\paragraph{Scene-level shot grouping.} This stage organizes cleaned shots into scenes and assigns stable identities to important subjects. We use Qwen3.5-122B-A10B~\citep{qwen2026qwen35} to inspect mosaics of representative frames from overlapping windows of 48 consecutive shots in a long video, with 12 shots shared between neighboring windows. Shots are grouped only when they share a physical place and a continuous event or narrative context. The presence of the same person or a similar color palette is not enough by itself. Although the large mosaics provide the model with the narrative context of the full video, we find that some local shots, particularly montage inserts, can still be assigned incorrectly. We therefore refine the initial scene plan in local windows of up to eight numbered frames; this correction pass moves shots only when the visual evidence is clear. A final rule requires every shot in a clip of at most one minute to belong to exactly one scene.

\paragraph{Subject discovery and linking.} We first identify the principal subjects in each shot and then match them across all shots in the same clip, enabling subject-consistent multishot captions. In our comparisons, directly using a multimodal model with explicit object-localization capabilities was more accurate for this task than assembling a complex pipeline of specialist models, such as the annotation pipeline used in MultiShotMaster~\citep{wang2026multishotmaster}. We therefore use Qwen3.5-122B-A10B for both subject discovery and cross-shot linking. For each shot, the model identifies up to six reusable subjects, including people, animals, vehicles, machines, and important objects, and returns a bounding box in the most representative frame for each subject. We describe the two-stage localization procedure in the next paragraph. The model then jointly examines the full scene-grouped clip, representative frames from every shot, and representative crops of the discovered subjects. The full clip provides evidence about actions and narrative continuity, while the selected frames provide evidence about appearance. Cross-shot matching relies on stable cues such as the face, clothing, shape, color, and material, rather than on a subject's temporary action or location. These stable subject identifiers connect the scene-level description with the description of each shot. We retain up to six reusable subjects for each clip. Events involving other visible subjects discovered at the shot level remain in the captions even when those subjects are not selected as reusable references.

\paragraph{Subject localization.} We localize each subject in two stages. A first pass uses the full scene context of a shot to choose a clear frame and a rough region. A second pass examines that exact frame and refines the bounding box.

\subsection{Retrieval and curation}

This stage selects clips that contain sustained, understandable events and removes duplicates. We retrieve candidates from coarse action captions and physical metadata, apply a fast story and motion screen with Qwen v4 Flash, and then review the complete video with Qwen3.5-122B-A10B. Editing cuts, flashing lights, subtitles, and camera shake can all produce high frame differences without useful subject motion, so optical flow, sharpness, brightness, and frame-change statistics only prioritize candidates and never decide quality on their own.

\paragraph{Content acceptance.} A strong candidate has recognizable subjects, clear visual quality, sustained motion, and an event with observable development, such as preparation, action, and outcome. A short attractive moment cannot compensate for a clip that is otherwise static, blurred, corrupted, or difficult to understand. The catalog records strong action, weaker but usable action, reviewed rejection, and not-yet-screened content as separate states; unscreened content is never counted as rejected.

\paragraph{Deduplication and balance.} We use the global clip captions produced in the preceding stage to support content review. We compare source-video identifiers, media signatures, time ranges, and caption signatures to detect duplicates. In most cases, only one clip is retained from a source video. A second clip is allowed for a rare topic only when its time range and shots do not overlap the first. We balance live action and animation and retain varied examples of human activity, machines and vehicles, groups, nonhuman creatures, and effects-rich environments. Once selected, a clip keeps its source mapping and selection reason, so a different file with the same name cannot silently replace it.

\subsection{Refinement and assets}

This stage converts the selected clips and annotations into visual, pose, and audio references for training our conditional generation model and improving its video-editing capabilities.

\subsubsection{Subject and background references.} 
We also use Qwen3.5-122B-A10B to perform a second screening of the subject and background assets. A subject reference should show stable identity features with little occlusion and enough of the subject visible to recognize it. If no suitable frame exists, the reference is marked unavailable rather than replaced with a poor crop. A background reference instead aims to show the layout of the scene. Foreground boxes and, when needed, object masks are combined before Big-LaMA~\citep{suvorov2022lama} fills the covered region. The completed background is accepted only after checking for remaining foreground content, damaged structure, and obvious texture artifacts.

\subsubsection{Pose tracking and asset states.}
We use Detectron2~\citep{wu2019detectron2} with a ViTDet backbone~\citep{li2022vitdet} to detect people and track their poses. People are detected in each frame and linked over time using box overlap, center movement, and changes in scale. Whole-body pose estimation then produces body and hand keypoints for the corresponding frames, while relative positions are preserved when several people appear together. A confirmed absence of people is recorded separately from a failed detector. Every asset is marked as available, not applicable, or blocked, which prevents missing outputs from being mistaken for valid ones.

\subsubsection{Audio alignment.} 
We use MOSS-Transcribe-Diarize 0.9B~\citep{yu2026mosstranscribe} to transcribe speech, separate speakers, and estimate utterance times. A speaker cluster is linked to a visible person only when presence, mouth movement, and timing support the match. Dialogue keeps its original language and word order. We then use Qwen3-Omni-Thinking~\citep{xu2025qwen3omni} to identify environmental sounds, contact sounds, nonverbal vocalizations, music within the scene, and background music, which are stored separately. This separation prevents the system from adding an expected sound merely because the corresponding action is visible.

\subsection{High-quality multishot data}

This final stage turns the selected clip and its coarse annotations into a detailed, evidence-based audiovisual description. The annotator reviews the video of no more than one minute together with its shot timeline, coarse captions, stable subject identities, and audio evidence. It describes composition, appearance, position, visible actions, state changes, camera motion, dialogue, and other sounds.

\paragraph{Dense audiovisual captions and validation.} The output contains subject definitions, an overall summary, ordered shot descriptions, the soundscape, background music, and links to the prepared assets. Automatic checks require continuous shot numbering, increasing cut times within the video duration, valid boxes, defined subject references, ordered audio events, nonempty required fields, and matching input signatures. A truncated structured response may be repaired only at the formatting level; the repair step is not allowed to invent visual or audio facts. Any factual, temporal, or identity conflict returns the sample for review.

\paragraph{Condition assets and division of labor.} Overall, our data pipeline follows a simple division of labor. Specialized models provide measurable evidence for cuts, text regions, speech times, and human poses. The multimodal model uses the full context to organize scenes, identities, events, and captions. Deterministic checks then ensure that all outputs refer to the same video, timeline, and subjects. This division makes the pipeline easier to audit and limits the spread of errors from one stage to the next.

\subsection{Video Editing Dataset}
Beyond general video data, we further construct and collect a large-scale video editing dataset. Training on this dataset substantially improves the model's ability to follow textual instructions and perform text-conditioned control.

\section{Method}
\subsection{Overview}
LynnReal-Omni is a native multimodal video generation framework built on a shared multimodal diffusion transformer. It unifies text-to-video, image-conditioned generation, reference-guided generation, structural control, editing, and long-video generation within a single model, and accepts heterogeneous visual inputs such as appearance references, editable 3D renders, and game recordings. The method consists of five main components. First, a native multimodal backbone with task-specific representation provides a shared denoiser and unambiguous conditioning interfaces. Second, multitask flow training and few-step distillation enable efficient generation across tasks, with a Flash variant that reduces denoiser depth and token count. Third, long-video generation is supported through a fixed head-overlap interface and bounded history storage. Fourth, a lightweight decoder is distilled to shift the inference bottleneck away from decoding. Fifth, agent-generated geometry and executable game controls supply inspectable spatial and motion guidance. Together, these components enable LynnReal-Omni to jointly generate video from heterogeneous multimodal conditions, producing coherent appearance, motion, interactions, and, when applicable, synchronized audio within a unified model. We describe each component below.

\subsection{Native Multimodal Backbone and Task representation}
The standard backbone contains 50 transformer blocks. Its residual width is 5,376, with 56 attention heads of dimension 128 and a feed-forward width of 14,336. Video latents have 24 channels and use $1\times2\times2$ patches. The audio stream has 32 input channels; text conditioning has dimension 5,120. The input/output projections and normalization-sensitive computations preserve their native precision conventions.

Each task packs modality tags, modality-specific row indices, 3D rotary positions and per-row noise times. Text and visual context are encoded before denoising. The denoiser then predicts the video and, when present, audio velocity in one forward. The video decoder reconstructs RGB frames from the denoised video rows; the audio decoder reconstructs the soundtrack separately.

\begin{figure}[t]
\centering
\includegraphics[width=\textwidth]{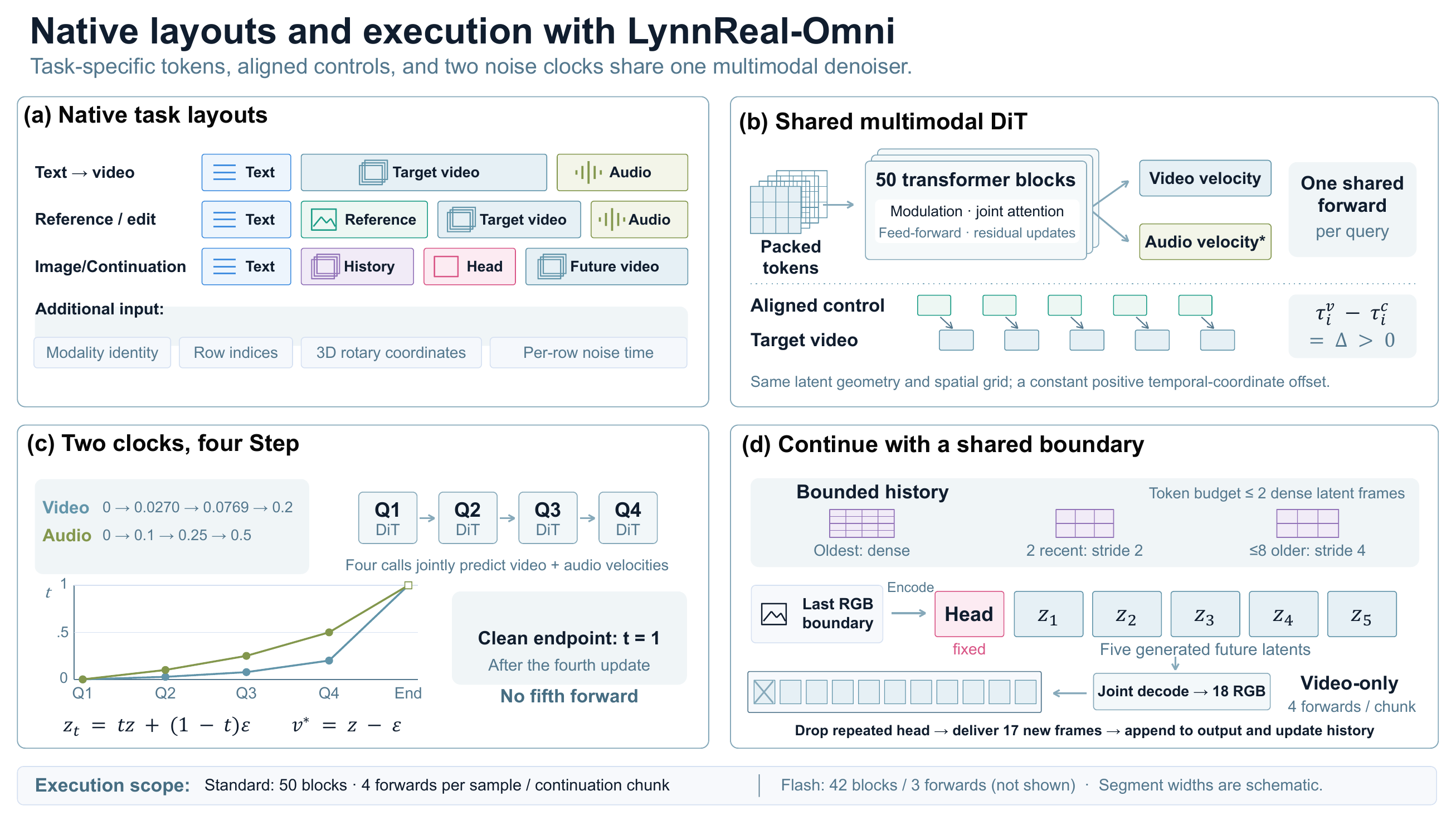}
\caption{Native layouts and execution. (a) Task-specific representation. (b) A shared denoiser preserves a positive temporal offset between aligned controls and target rows. (c) Video and audio have different clean-time coordinates within the same four step. (d) long video generation jointly decodes one fixed head and five future latents, then removes the repeated boundary frame to deliver 17 new RGB frames. Audio participation follows the task's training contract.}
\label{fig:dit}
\end{figure}

Image conditioning has two explicit interfaces. In the unified reference layout, supplied images occupy ordered picture-reference segments and an opening-frame instruction is semantic. The native keyframe layout instead encodes first and optional last images into the target's keyframe partition. The release selects this layout through the model bundle's conditioning contract; it is available to the full standard DiT as well as Flash. A native 768p standard-model check animates a harbor photograph while retaining the main scene layout. This interface distinction is not a guarantee of pixel-exact image reconstruction, and prompt formatting follows the selected contract.

A semantic video reference may retain its own resolution and provide appearance or motion cues. A pose sequence, editing source, depth sequence, or game render instead describes the target timeline frame by frame. These controls are sampled on the 24-fps model clock, resized onto the target canvas, and extended with their last valid frame if shorter than the requested target. Source frame rate, selected duration and any terminal padding must therefore be recorded; a padded tail is not new observed motion. The reference and target then have equal latent geometry and identical spatial grids. Their temporal positions satisfy
\begin{equation}
  \tau^{\mathrm{target}}_i-\tau^{\mathrm{control}}_i=\Delta,
  \qquad \Delta>0,
\end{equation}
with a constant offset for every corresponding latent frame. The positive offset keeps reference and generated tokens in separate domains; setting their absolute rotary coordinates equal changes the model's conditioning semantics.

\subsection{Multitask Training and Few-Step Distillation}
Let $z$ be a clean target latent and $\epsilon\sim\mathcal{N}(0,I)$. We use the clean-time convention
\begin{equation}
  z_t=t z+(1-t)\epsilon,\qquad v^\star=z-\epsilon.
\end{equation}
The denoiser learns the velocity on the task's valid target rows. Conditioning rows and padded target rows are excluded from the corresponding target loss. Task sampling supplies text generation, image conditioning, reference controls, editing, and long video generation to the shared model. Dataset windows, caption text, modality masks, and temporal alignment are carried together so that visual controls cannot silently refer to a different interval from the target.

The video noise level is drawn from a shifted logistic-normal distribution. For \(g \sim \mathcal{N}(0,1)\) and \(u = \operatorname{Sigmoid}(g)\), we define the video noise level
\begin{equation}
 s_v = \frac{3u}{1+2u} \in [0,1],
 \qquad t_v = 1 - s_v,
\end{equation}
where \(t_v\) denotes the corresponding clean time. This reparameterizes the time coordinate from noise level to clean time; the velocity target \(v^\star = z - \epsilon\) remains unchanged. When valid target audio exists, its noise level follows the backbone's aligned audio clock. We first map \(s_v\) to an intermediate level \(s_b = s_v/(12 - 11 s_v)\), and then obtain the audio noise level
\begin{equation}
 s_a = \frac{3s_b}{1+2s_b},
 \qquad t_a = 1 - s_a.
\end{equation}
This aligned clock ensures that video and audio are noised according to their respective schedules while sharing the same random draw \(g\).
Video and audio each use independently sampled Gaussian noise. The objective averages the squared velocity error over valid entries in each modality,
\begin{equation}
 \mathcal{L}=\operatorname{MSE}_{\mathcal{V}}(\widetilde v_v,z_v-\epsilon_v)
 +0.1\,I_{\mathrm{valid\ audio}}\operatorname{MSE}_{\mathcal{A}}(\widetilde v_a,z_a-\epsilon_a).
\end{equation}
Missing audio is not treated as a supervised silent recording.

The joint model also uses guidance-aware fitting. The same denoiser supplies a captionless, stop-gradient prediction $v_\varnothing$ while retaining the supplied visual controls. For the recorded training scale $w=3$, the loss above uses
\begin{equation}
 \widetilde v=\frac{v_c+(w-1)\operatorname{sg}(v_\varnothing)}{w}.
\end{equation}
Only the conditional branch receives gradients. At its regression optimum this encourages $v_c\approx wv^\star-(w-1)v_\varnothing$. Inference uses the conditional prediction directly; it does not add an unconditional forward to each denoising step. Cached target video latents use the posterior mode, while reference encoding retains its native posterior convention.

The standard model retains the full backbone and adopts low-rank trajectory distribution matching~\citep{luo2025tdm} for four-evaluation inference. Training proceeds over four non-overlapping transition intervals, differentiating through one selected student transition. A frozen teacher and a separately trained full-depth fake-score critic evaluate noise conditioned on the student endpoint. The critic is trained with importance-weighted clean-target regression under clipped signal-to-noise weighting, while the student uses the normalized pseudo-Huber surrogate described in Section~\ref{sec:flash}. To improve the dynamics of distribution matching distillation (DMD), we incorporate the dynamic reward mechanism from Reward Forcing~\citep{lu2026reward}. This visual reward reweights the video term of the combined video--audio surrogate, whereas the audio term receives a fixed weight because the reward model is not audio-aware. Deployment uses the student's exponential moving average and requires neither the teacher nor the critic.


Although video and audio share the same four denoising evaluations, they are conditioned on distinct modality-specific noise schedules. Under the native 768P configuration, the video branch is evaluated at clean-time timesteps of approximately $(0,0.0270,0.0769,0.2000)$, whereas the corresponding audio timesteps are $(0,0.1000,0.2500,0.5000)$. These timestep values are determined by the modality-specific row-to-time mappings, rather than by directly indexing a shared timestep vector, since the same denoising evaluation can correspond to different noise levels for the two modalities. After the fourth denoising update, both modalities advance to their clean endpoints without requiring an additional network evaluation. Figure~\ref{fig:dit} illustrates how these four shared denoising evaluations are realized within the packed multimodal layout.

\subsubsection{Flash variant and inference acceleration.}\label{sec:flash}


LynnReal-Omni-Flash accelerates text-to-video and native keyframe-conditioned generation through transformer depth reduction, spatial token compression, and three-step distillation. The model retains 42 of the 50 transformer blocks and requires only three denoising evaluations per sample.

To reduce intermediate computation while preserving multimodal conditioning, the first two transformer blocks operate on the full token sequence, enabling early integration of video, text, and audio features. The subsequent 26 blocks employ frame-wise spatial token compression: video tokens are subsampled with a stride of two along both spatial dimensions, reducing the spatial token count to approximately one quarter, while retaining boundary rows and columns to preserve edge information. Text and audio tokens remain uncompressed, and all retained video tokens preserve their original spatiotemporal rotary positional coordinates. This design reduces the dominant spatial computation without sacrificing temporal resolution or altering the multimodal token structure.

We preserve the full-resolution features through a residual connection. Let $H$ denote the features entering the compressed stack, $P$ the token-selection operator, and $F_{\mathrm{mid}}$ the middle blocks. Before the final fourteen full-resolution blocks, we reconstruct
  \begin{equation}
      H_{\mathrm{out}}
      = H + U\!\left(F_{\mathrm{mid}}(PH)-PH\right),
  \end{equation}
where $U$ assigns each omitted video token the feature update of its nearest retained spatial token in the same frame; retained tokens receive their own updates. This preserves the original full-resolution features while allowing the compressed stack to supply contextual updates. The final blocks then refine all tokens jointly. For $N_v$, $N_t$, and $N_a$ video, text, and audio tokens, the middle-stack sequence length decreases from $N_v+N_t+N_a$ to approximately $N_v/4+N_t+N_a$. This reduces both token-wise projection and feed-forward computation, as well as the sequence-length-dependent attention cost.

We train the reduced student using trajectory distribution matching~\citep{luo2025tdm}, with a frozen full-depth teacher and a trainable full-depth fake-score critic. Each worker backpropagates through one selected transition of the student's three-step trajectory. Teacher and critic predictions are evaluated at the same noisy state, sampled within that transition's interval conditional on the student endpoint. The critic learns to predict generated clean targets using importance-weighted regression with clipped signal-to-noise weighting.

For the generator objective, let $z_G$ be the student clean prediction and $\hat z_F,\hat z_T$ the corresponding critic and teacher predictions. Define $d=\hat z_F-\hat z_T$ and
  $a=\max(\operatorname{mean}|z_G-\hat z_T|,10^{-6})$. We use
  \begin{equation}
   \mathcal{L}_{\mathrm{DM}}
   =
   \frac{
   \operatorname{mean}\!\left[
   \rho_c\!\left(z_G-\operatorname{sg}(z_G-d)\right)
   \right]}
   {\operatorname{sg}(a)},
   \qquad
   \rho_c(e)=\sqrt{e^2+c^2}-c,\quad c=10^{-3}.
  \end{equation}
Here $\operatorname{sg}$ denotes stop-gradient, and normalization is applied outside the robust penalty. Visual reward reweights only the video loss; the audio loss has a fixed weight. We additionally use real-data flow regression on aligned video and valid audio. Audio interval supervision matches the student's update to the teacher's finer integration over the same interval, using an RMS-normalized smooth-$L_1$ loss with a normalization floor and separate validity masks for the stereo channels. These auxiliary computations are used only during training; inference requires three student evaluations followed by codec decoding.

The final Flash DiT adopts W4A8 quantization—INT4 weights with FP8 activations—to further reduce memory traffic and accelerate matrix multiplications, while precision-sensitive modules remain at higher precision. Operator fusion lowers intermediate memory accesses and kernel-launch overhead. We evaluate these optimizations independently of depth reduction and token compression to isolate their effects on latency and generation quality.

\subsection{Long Video Generation with Bounded History}




\subsubsection{Chunk-wise video generation}

Long videos are generated in fixed-length chunks. At each step, the model conditions on the preceding video context and generates 17 new frames. To maintain temporal coherence across adjacent chunks, the final RGB frame of the preceding chunk is reused as a shared boundary frame. Its latent remains fixed while the model predicts 5 future latent frames, which are jointly decoded with the boundary latent; the duplicated boundary frame is then removed. This design provides a consistent transition across chunks while preserving a fixed generation window. Each chunk requires 4 denoiser evaluations, without frame interpolation or exposure correction.

\subsubsection{Compact temporal context}

We construct a fixed-budget temporal context that preserves fine-grained recent dynamics while retaining coarse long-range information. Specifically, the earliest latent frame is kept at full spatial resolution, 2 recent latent frames are sampled with spatial stride 2, and up to 8 earlier frames are sampled with stride 4. The shared boundary frame is excluded to avoid redundant conditioning. Across the evaluated resolutions, the resulting context contains no more tokens than 2 full-resolution latent frames, keeping attention cost independent of video length.

The temporal context is also bounded in storage: once the capacity is reached, the initial frame and the most recent frames are retained while intermediate entries are discarded. Together, bounded storage and fixed-budget tokenization enable long-form generation without increasing the temporal conditioning cost.

\subsection{Lightweight Decoder Distillation}\label{sec:lightvae}

We distill the original video decoder into a shallower student to reduce decoding cost while preserving the latent interface of the pretrained denoiser. Our approach combines teacher reconstruction, spatial and temporal supervision, and consistency through the frozen encoder. Paired source pixels provide additional detail supervision when a valid correspondence is available.

\begin{figure}[h!]
\centering
\includegraphics[width=\linewidth]{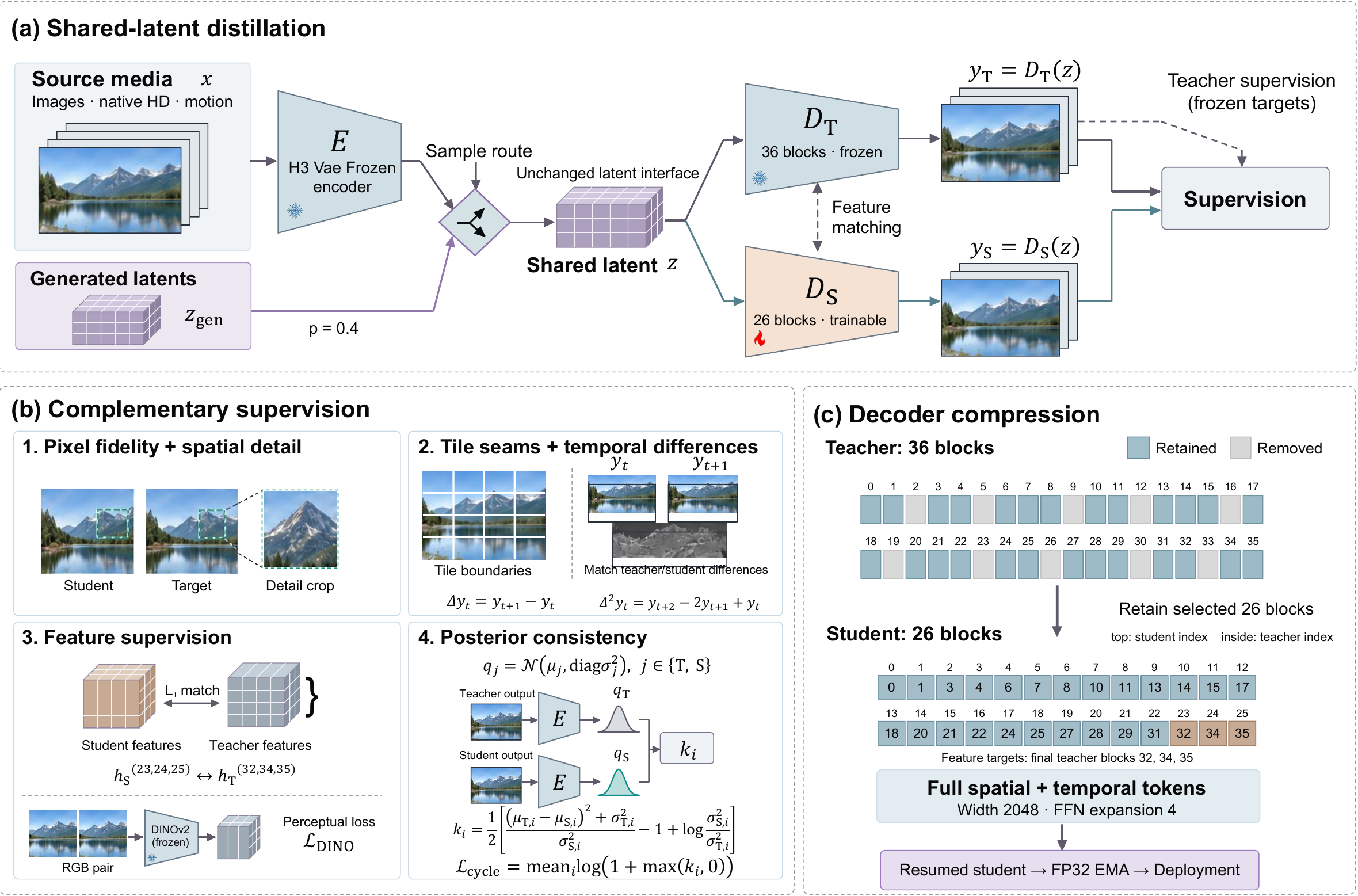}
\caption{Lightweight decoder distillation. The teacher and student decode the same latent. Pixel, temporal, and feature losses transfer the teacher's reconstruction behavior, while a frozen encoder constrains the distributions of their re-encoded outputs. Genuine paired source pixels provide additional detail supervision.}
\label{fig:vae_training}
\end{figure}

\subsubsection{Architecture and initialization}

The student decoder $D_S$ is obtained by reducing the depth of the 36-block teacher decoder $D_T$ to 26 transformer blocks while preserving the hidden dimension of 2,048 and an MLP expansion ratio of 4. Each retained student block is initialized from its corresponding teacher block, which also establishes the block-wise correspondence used for feature distillation. The retained teacher blocks are
\begin{equation}
\begin{aligned}
\mathcal{I}=\{&
0,1,3,4,6,7,8,10,11,13,14,15,17,\\
&18,20,21,22,24,25,27,28,29,31,32,34,35
\}.
\end{aligned}
\end{equation}

The latent representation is left unchanged: the encoder, latent dimensionality, spatial scaling, and temporal structure are identical to those of the teacher model, and no spatial or temporal latent tokens are removed during decoding. The distilled decoder can therefore replace $D_T$ directly without modifying the upstream generative model or re-encoding its latent outputs. The final decoder is obtained by further refinement of this initialized student.

\subsubsection{Training distribution and supervision}

Let $E$ denote the frozen encoder. To expose $D_S$ to both encoded data latents and the latent distribution encountered during generation, training samples are drawn from 2 sources. With probability 0.6, we use latents $z=E(x)$ encoded from the training data; with probability 0.4, we use latents produced by the generative model. Within the data-encoded branch, native-HD and high-motion samples are each selected with probability 0.2. Native-HD samples have a short-side resolution of at least 720 pixels, while videos recorded at $\geq 20$ fps are densely sampled with probability 0.7. Each clip contains 2--33 frames under a budget of 4 million pixel-frames.

The supervision target is determined by the latent source. For a latent $z$, the default target is the frozen teacher reconstruction
\begin{equation}
y_T = D_T(z).
\end{equation}
Model-generated latents are supervised exclusively by $y_T$, as they have no paired pixel-space target. We further improve robustness to deviations from the encoder latent distribution by perturbing latents with probability 0.25 using Gaussian noise with a relative standard deviation of 0.03; these perturbed latents are likewise supervised by $D_T$. Single-frame latent slices are sampled with probability 0.1 to strengthen image decoding.

For image samples, we additionally use a frozen image-specialized decoder $D_I$. Generated image latents are supervised by $D_I$. For encoded training images, the teacher is selected according to reconstruction fidelity: $D_I$ replaces $D_T$ when its clamped reconstruction yields a lower pixel-space $L_1$ error with respect to the input image. This adaptive teacher selection exploits the stronger image reconstruction capability of $D_I$ without sacrificing fidelity to the original data.

\subsubsection{Distillation objectives}

Given a latent $z$, the student reconstruction is $y_S=D_S(z)$. Training combines teacher reconstruction, structural regularization, feature distillation, perceptual supervision, latent consistency, and, when paired data are available, direct pixel-space reconstruction:
\begin{equation}
\begin{aligned}
\mathcal{L} ={}&
\alpha \mathcal{L}_{\mathrm{rec}}
+\mathcal{L}_{\mathrm{reg}}
+\lambda_{\mathrm{feat}}\mathcal{L}_{\mathrm{feat}}
+\lambda_{\mathrm{perc}}\mathcal{L}_{\mathrm{perc}}\\
&+m_{\mathrm{cyc}}\lambda_{\mathrm{cyc}}\mathcal{L}_{\mathrm{cycle}}
+\lambda_{\mathrm{src}}\mathcal{L}_{\mathrm{src}}
+m_{\mathrm{img}}\lambda_{\mathrm{LPIPS}}\mathcal{L}_{\mathrm{LPIPS}}.
\end{aligned}
\end{equation}

The primary distillation term matches the student output to the selected teacher reconstruction:
\begin{equation}
\mathcal{L}_{\mathrm{rec}}
=
\operatorname{mean}\left|y_S-y_T\right|
+
5\,\operatorname{mean}\left(y_S-y_T\right)^2.
\end{equation}
The $L_1$ component preserves reconstruction accuracy, while the quadratic term places stronger emphasis on large pixel deviations. The regularization term
\begin{equation}
\mathcal{L}_{\mathrm{reg}}
=
\sum_k \lambda_k \mathcal{R}_k
\end{equation}
captures complementary spatial and temporal reconstruction properties. Multiscale, gradient, high-pass, wavelet, and SSIM terms constrain appearance, edges, and local detail; patch- and tile-boundary losses suppress decoding seams; temporal first- and second-order differences regularize frame-to-frame motion and acceleration; and spatial- and temporal-phase terms reduce periodic reconstruction artifacts. Their default weights are summarized in Table~\ref{tab:vae_losses}.

\begin{table}[t]
\centering
\small
\caption{Default auxiliary loss weights. Source-repair samples use the target and coefficient adjustments described in the text.}
\label{tab:vae_losses}
\begin{tabular}{lr@{\qquad}lr}
\toprule
Spatial loss & Weight & Temporal or feature loss & Weight \\
\midrule
Multiscale pyramid & 0.25 & Temporal difference & 0.20 \\
Spatial gradient & 0.10 & Temporal acceleration & 0.05 \\
High-pass residual & 0.15 & Temporal phase & 0.10 \\
Haar wavelet bands & 0.10 & Decoder features & 0.20 \\
Wavelet energy & 0.02 & DINOv2 features & 0.05 \\
Patch boundary & 0.20 & Encoder consistency & 0.01 \\
Tile boundary & 1.50 & & \\
Spatial phase & 0.05 & & \\
SSIM & 0.10 & & \\
\bottomrule
\end{tabular}
\end{table}

Feature distillation further aligns the internal representations of the student and teacher. We apply an $L_1$ loss to the final 3 student blocks and their corresponding teacher blocks, assigning a $4\times$ weight to the final block. To reduce training cost, this loss is evaluated on a reduced latent region with spatial size 16 and at most 4 temporal tokens. In parallel, frozen DINOv2 ViT-S/14 features~\citep{oquab2023dinov2}, extracted at 224-pixel resolution, provide the perceptual objective $\mathcal{L}_{\mathrm{perc}}$.

When paired training pixels are available, we additionally supervise the student directly with the input $x$ using a Charbonnier loss:
\begin{equation}
\mathcal{L}_{\mathrm{src}}
=
\operatorname{mean}
\sqrt{
\left(y_S-x\right)^2+\epsilon^2
}.
\end{equation}
This source-reconstruction branch is applied to eligible image and low-resolution video samples. For these samples, we use $\alpha=0.25$ and $\lambda_{\mathrm{src}}=2$, and the original pixels also serve as targets for the applicable detail losses. Image reconstruction additionally enables VGG LPIPS~\citep{zhang2018perceptual} with weight 0.1 while disabling the wavelet-energy term. Outside this branch, $\alpha=1$. For encoded images assigned to the specialized image teacher, $\lambda_{\mathrm{src}}=1$; otherwise, $\lambda_{\mathrm{src}}=0$. Consequently, model-generated and perturbed latents receive teacher supervision without direct pixel-space reconstruction.

We further regularize the student in latent space by requiring teacher and student reconstructions to produce consistent encoder posteriors. A shared contiguous window of at most 5 frames is sampled from $y_T$ and $y_S$, clamped, resized to 256 pixels, and re-encoded by $E$. Let
\begin{equation}
q_T
=
\mathcal{N}
\left(
\mu_T,
\operatorname{diag}\sigma_T^2
\right),
\qquad
q_S
=
\mathcal{N}
\left(
\mu_S,
\operatorname{diag}\sigma_S^2
\right).
\end{equation}

For each posterior coordinate, we compute the KL divergence
\begin{equation}
k_i
=
\frac{1}{2}
\left[
\frac{
\left(\mu_{T,i}-\mu_{S,i}\right)^2+\sigma_{T,i}^2
}{
\sigma_{S,i}^2
}
-1
+
\log
\frac{
\sigma_{S,i}^2
}{
\sigma_{T,i}^2
}
\right],
\end{equation}
and define
\begin{equation}
\mathcal{L}_{\mathrm{cycle}}
=
\operatorname{mean}_i
\log\!\left(
1+\max(k_i,0)
\right).
\end{equation}

The logarithmic transformation limits the contribution of large divergences while preserving their gradients. Log variances are clamped to $[-20,10]$ for numerical stability. The teacher posterior is detached, whereas gradients are propagated through the frozen encoder to update the student reconstruction.

Importantly, $\mathcal{L}_{\mathrm{cycle}}$ aligns the posterior of the student reconstruction with that of the teacher reconstruction, rather than directly matching the original input latent. The loss is activated after 800 training steps, evaluated every 8 updates, and disabled for source-reconstruction samples, for which the original pixels already provide direct supervision.
```




\subsubsection{Optimization and deployment}

The refinement stage is optimized with AdamW using a learning rate of $3\times10^{-6}$, betas $(0.9,0.95)$, $\epsilon=10^{-8}$, zero weight decay, and gradient clipping at norm 1. Each worker accumulates gradients over 2 single-sample microbatches. Model parameters and accumulated gradients are maintained in FP32, while forward computation uses BF16 autocast. Training resumes with a 100-step learning-rate warmup.

To control memory usage during decoder training, videos are processed with rectangular tiles of $272\times208$ pixels, using overlaps of 0 and 16 pixels along the respective axes and a tile batch size of 2. Images use $256\times256$ tiles with a repeated 5-latent context and output phase 3. At inference time, we retain the original decoder tiling configuration by default, while adaptive tiling is optionally enabled for different memory--latency trade-offs. Both settings preserve the native temporal padding and output trimming.

Checkpoints are selected using a fixed reconstruction benchmark that jointly evaluates spatial fidelity and temporal consistency. Decoder depth, numerical precision, and tiling strategy are additionally ablated independently to isolate their effects on reconstruction quality and inference latency.

\subsection{Temporal repair of generated-video artifacts}
\label{sec:video-repair}
\method{} supports instruction-guided repair of generated-video artifacts through its shared image-editing interface. Given a corrupted video and a repair instruction, the model processes each source frame and reassembles the repaired images in their original temporal order. The instruction specifies which artifacts to suppress and which scene content to retain. This procedure uses the same standard DiT across frames and does not require a scene-specific restoration network. The implementation accepts an external video and prompt; the candle sequence below is an illustrative example rather than a restriction of the repair interface.

\paragraph{General framewise repair.}
For a source sequence $\{x\}$ and instruction $p$, we independently edit each $x_t$ with the shared model, using the same seed across frames, and retain one generated image per source timestamp. The current implementation produces a short internal clip for each image edit and selects a configurable frame from that clip. This accommodates the video decoder without adding frames to the source timeline. Optional appearance references or temporally aligned guide videos can also be supplied. Each source-frame edit uses \textbf{four denoiser evaluations}, giving a total sampling cost of $4T$ evaluations for $T$ source frames. Temporal repair here denotes restoration across the complete sequence with its frame order and playback timing retained; the frame edits are independent and impose no explicit cross-frame consistency constraint.

\begin{figure}[htbp]
\centering
\includegraphics[width=\linewidth]{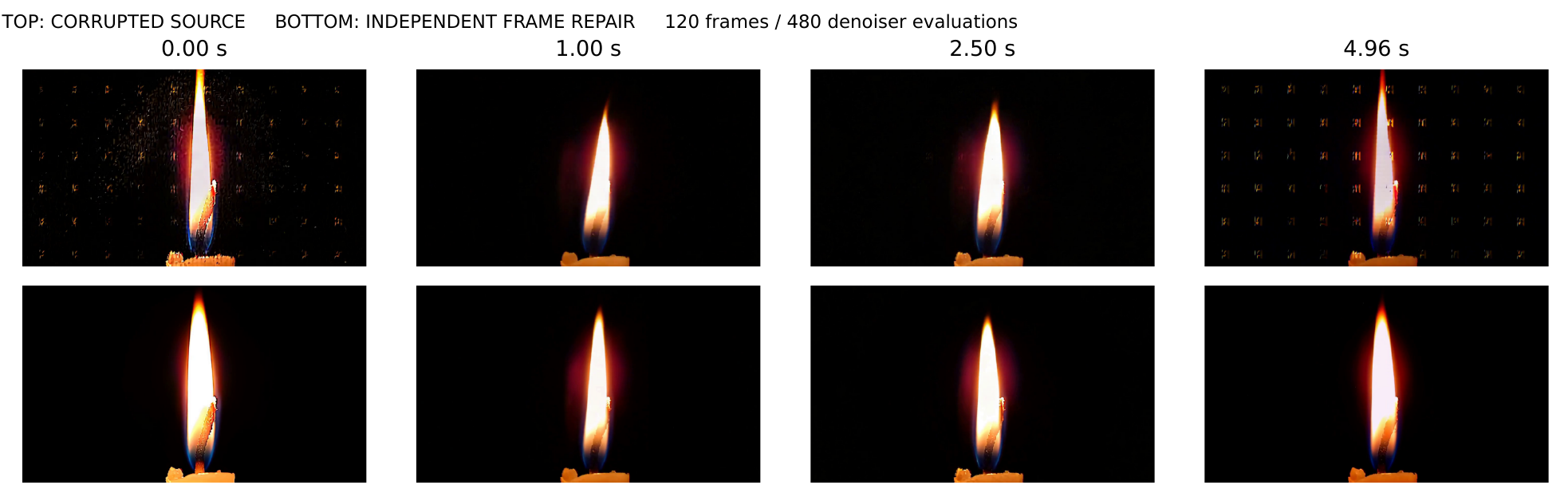}
\caption{An example of generated-video artifact repair with \method{}. Top: corrupted candle video. Bottom: independently repaired frames at identical timestamps, including both endpoints. Full frames are displayed at the same scale. Each source-frame edit uses four denoiser evaluations, for 480 evaluations across this 120-frame example. The accompanying demo plays both complete videos synchronously.}
\label{fig:candle-repair}
\end{figure}

\paragraph{Illustrative result.}
We apply this general procedure to a five-second candle video containing bright speckles, background grid patterns, and vertical rendering artifacts. The example uses the standard BF16 DiT, seed 7, and a 1344$\times$768 output canvas. Its prompt requests artifact removal while retaining the observed candle and flame; its configuration selects the final frame (index 21) of each 22-frame internal clip. All 120 source frames are independently repaired, requiring \textbf{480 denoiser evaluations in total}. This selected run uses the raw source frames without an appearance reference, filtered guide, generated-frame feedback, temporal interpolation, or subsequent restoration filter.

\paragraph{Full-sequence assessment.}
Inspection of all 120 repaired frames shows substantially fewer background grid artifacts and speckles than the corrupted source and the preceding joint video-edit trial, while the flame continues changing over time. Small brightness and shape fluctuations, an occasional residual point, and reduced flame-height variation remain. Complete synchronized videos and fixed-mask flame-tip and background-bright-pixel traces accompany the example. These observations provide qualitative evidence for this unpaired case, without clean ground truth; they do not establish pixel-exact recovery or restoration quality across arbitrary scenes. Temporal consistency remains a limitation of independent frame editing.

\subsection{Agent-generated 3D scene and executable game controls}

We use agents to construct executable 3D scenes from reference images and task descriptions, providing spatial and motion guidance for subsequent video generation. The scene representation combines geometry, materials, object transforms, lighting, and cameras with explicit state updates and control inputs. Code execution, rendered observations, and iterative refinement connect these components, allowing appearance, motion, and interaction to be specified and revised within a shared editable representation.

\paragraph{Image Analysis and Scene Construction.}
Scene construction begins with an analysis of the reference image and task description. The image provides evidence for object silhouettes, relative scales, occlusions, and support relationships, while the task description specifies the required actions and interactions. These observations inform the environment, object models, appearance, and control logic. Shared coordinate conventions and interfaces support independent development and revision of each component.
Objects and their parts are represented by meshes and hierarchies, with transforms specifying position, orientation, and scale. Materials, procedural textures, and lighting define surface appearance. The agent implements the main structures and basic behavior before refining object shape and environmental detail, using execution feedback to identify and correct implementation errors. Absolute scale and hidden geometry are resolved through modeling assumptions consistent with the visible evidence.

\paragraph{Pose and Camera Refinement.}
Rendered views from the reference viewpoint guide the refinement of camera parameters and object transforms. Comparisons with the input image focus on framing, silhouettes, perspective, and occlusion, with each update evaluated through a new render. Correspondence fitting can assist parameter estimation within this process, while image feedback guides refinement across scenes.
Refinement proceeds from the overall viewpoint and object arrangement to part proportions, materials, and lighting, reducing the risk that local geometry changes compensate for an incorrect perspective. Assembly transformations preserve internal relationships, and pose updates are accompanied by renewed checks of contact and occlusion. Code or configuration snapshots and rendered views are retained to support comparison and revision. These comparisons provide construction feedback rather than an independent measure of recovered 3D accuracy.

\paragraph{Contact Correction and Scene Validation.}
Scene validation combines appearance assessment with checks of spatial consistency. Reference-view renders support comparisons of composition, silhouettes, and materials, while additional viewpoints reveal hidden geometry, support relationships, and intersections. Geometric measurements and collision checks localize gaps and overlaps for correction through object height, assembly transforms, or local geometry. Corrections are followed by checks of nearby objects to account for their effects on surrounding contacts and occlusions.
Runtime checks accompany visual refinement to verify that changes preserve object hierarchies, animation, and rendering performance. Shared geometry, instancing, and static merging reduce rendering cost where applicable. The resulting scene retains editable objects, materials, and camera settings and is validated through loading and inspection from multiple viewpoints. Runtime behavior, geometric validity, and visual fidelity are recorded as separate aspects of validation.

\paragraph{Game Logic and Temporal Control.}
Game logic defines motion and interaction through explicit state updates. The state includes object position, velocity, orientation, and action phase, while control inputs specify movement, turning, and action triggers. Update rules account for acceleration, gravity, ground contact, and collisions and produce the corresponding interaction events. Object poses, part animations, and cameras are updated from this state to generate visible motion. Automated control selects inputs according to the current state and task goals through the same update logic used for manual control. Initial conditions, control policies, and behavior parameters therefore provide direct means of revising the resulting motion.

The output sequence is determined jointly by the simulation and camera behavior. Cameras follow configured trajectories or tracking rules, and a recording script advances the simulation at fixed output time intervals, capturing each frame after rendering completes. States and interaction events are stored alongside the frames, linking visible motion to program execution and decoupling the time required to record the sequence from its playback speed. The encoded video is reviewed for appearance, motion continuity, and the intended actions, with the corresponding program version, control settings, and execution records retained. This rendered sequence supplies layout and motion references for subsequent video generation, while the underlying program supports further edits to appearance and behavior.

\section{Experiments}
\label{sec:experiments}
\subsection{Experimental questions and settings}
Our experiments address three distinct questions: whether the released sampler preserves its reference computation, which decoder and execution changes improve measured efficiency, and how visual control behaves over time. Table~\ref{tab:evaluation-settings} specifies the unit of evidence for each study. Within a paired comparison, we hold source media, prompt, seed and the relevant model artifact fixed. Between studies, changes in output geometry, decoder context or arithmetic are stated explicitly. This organization avoids treating a reconstruction measurement as a generated-video score, or transferring a small-canvas timing to a larger-canvas demonstration.
\begin{table}[!ht]
\centering
\small
\setlength{\tabcolsep}{4pt}
\renewcommand{\arraystretch}{1.16}
\begin{tabular}{@{}>{\raggedright\arraybackslash}p{0.19\linewidth}>{\raggedright\arraybackslash}p{0.27\linewidth}>{\raggedright\arraybackslash}p{0.45\linewidth}@{}}
\toprule
Study & Geometry and execution & Controlled comparison and interpretation \\
\midrule
Native inference & 1344$\times$768; 124 frames; four evaluations; full decoder & Fixed prompts, seeds, checkpoint and adapter bytes. Tests release/reference fidelity and visible event structure. \\
Decoder reconstruction & 1344$\times$768; one image, 22- and 124-frame clips & Shared official-encoder latent; matched image context. Isolates decoder depth and tile scheduling; metrics precede encoding. \\
Warm latency & 960$\times$544 native canvas, cropped to 540p; 22 frames & Synchronized repeated timings, two warmups, five retained calls. Codec, attention and precision are stated per row. \\
Continuation & 720 new frames; 43 chunks; four evaluations per chunk & Bounded stored history and observed memory; paired scene/seed controls. Visual drift is assessed separately from memory growth. \\
MSAVP benchmark & 100 prompts; 20 reported metrics; 720p, 15-s outputs & Cross-model scores are reported separately for T2V and I2V.  \\
\bottomrule
\end{tabular}

\caption{Evaluation settings and the question supported by each experiment. Individual artifacts retain the full invocation and hashes. Development cases are reused for diagnosis and are not a held-out estimate of general quality.}
\label{tab:evaluation-settings}
\end{table}

\subsection{MSAVP: complex audiovisual and physical evaluation}
\label{sec:msavp}

\subsubsection{Benchmark overview}

\paragraph{MultiShot-AV-Physics Bench.}
We introduce the MultiShot-AV-Physics Bench (MSAVP) for multishot audiovisual generation, a setting covered by relatively few existing benchmarks. MSAVP emphasizes instruction following, cross-shot consistency, physical plausibility, and audiovisual coordination. Its 20 reported metrics are summarized in Table~\ref{tab:msavp-metrics}; the quality criteria draw on WBench~\citep{ying2026wbench}, while the physics criteria are informed by VideoPhy~\citep{bansal2024videophy} and PhyGDPO~\citep{cai2026phygdpo}. Unlike conventional evaluation pipelines, MSAVP employs an agentic VLM to plan and execute the assessment, enabling structured semantic reasoning over complex requests. A fixed checklist for each prompt constrains this reasoning to explicit, verifiable conditions, making the evaluation more consistent and auditable.

\paragraph{Prompt construction.}
The benchmark tests dense events and cross-shot consistency. We do not adapt prompts to the outputs of any evaluated model. We constructed 100 prompts through text review and deterministic coverage search, using a common format for all systems. Each prompt contains an integrated audiovisual description, an overall soundscape, and an optional description of non-diegetic music. Twenty-two prompts request music, while the remaining 78 do not. The set includes 54 single-shot prompts and 46 multi-shot prompts: 10 request two shots, 7 request three, 9 request four, 16 request five, and 4 request six.

The prompts also cover a broad range of content rather than concentrating on a single failure mode. They span 99 overlapping topic or capability tags and 95 distinct primary topics, including daily activities, sports, machinery, material deformation, natural events, fantasy, science fiction, and demanding camera work.

\subsubsection{Evaluation workflow}

\paragraph{Inputs and generation settings.}
T2V and I2V receive identical text prompts. In I2V, every model also receives the same 1280$\times$720 RGB opening image for a given prompt. Each system produces a 15-second, 16:9 video at 1280$\times$720 resolution and 24 fps. Seedance 2.0 is evaluated only in T2V because the official API's strict moderation of photorealistic human input images prevented completion of a comparable 100-prompt I2V run.

For the locally executed systems, LynnReal-Omni uses our FL2VA checkpoint for both T2V and I2V. LTX-2.5 uses the official DFR pipeline with BF16 weights, runtime FP8 casting, CPU offloading, chunked-eager DiffVAE decoding, and one spatial upsampling stage on one 80-GB H100. It generates 361 frames at 1280$\times$704 and 24 fps, with 8-pixel vertical padding on each side for delivery at 1280$\times$720. Cosmos3-Super uses the complete non-distilled BF16 I2V model for 35 steps with CFG 6 and shift 10, following the official four-H100 Ray/FSDP configuration with diffusion caching, \texttt{torch.compile}, and CUDA Graphs. MiniMax-H3 uses the complete unquantized SGLang \texttt{fl2va} model without Turbo LoRA, running 50 steps with flow shift 12, audio flow shift 3, tensor parallelism 2, Ulysses parallelism 2, and the speed performance mode on four 80-GB H100s. Its 15-second 16:9 output is generated at a short side of 768 pixels and proportionally resized or padded to the common 1280$\times$720, 24-fps delivery format.

\begin{table*}[!h]
\centering
\caption{MSAVP family and overall scores. Metrics are averaged within each capability family, followed by an equal average of the six family scores.}
\label{tab:msavp-overview}
\small
\setlength{\tabcolsep}{4.5pt}
\renewcommand{\arraystretch}{1.08}
\begin{tabular}{@{}llrrrrrrr@{}}
\toprule
Mode & Model & Prompt & Event & Visual & Multi-shot & World & Audio & Overall \\
\midrule
T2V & Seedance 2.0  & \textbf{83.79} & \textbf{74.62} & \textbf{61.53} & 84.50 & 87.87 & 84.01 & \textbf{79.39} \\
    & MiniMax-H3-FL2V    & 82.06 & 73.52 & 59.13 & \textbf{86.75} & \textbf{88.79} & \textbf{85.35} & 79.27 \\
    & LynnReal-Omni & 83.69 & 72.32 & 59.01 & 83.25 & 86.00 & 82.26 & 77.76 \\
    & LTX-2.5       & 72.02 & 59.51 & 54.52 & 30.08 & 74.13 & 73.07 & 60.56 \\
    & Cosmos3-Super & 65.42 & 52.64 & 55.79 & 5.76 & 74.72 & 71.25 & 54.26 \\
\addlinespace[2pt]
I2V & MiniMax-H3-FL2V    & 85.16 & 74.44 & \textbf{61.53} & \textbf{87.36} & \textbf{91.32} & \textbf{84.11} & \textbf{80.65} \\
    & LynnReal-Omni & \textbf{86.55} & \textbf{74.73} & 61.16 & 81.81 & 89.52 & 81.44 & 79.20 \\
    & LTX-2.5       & 77.31 & 61.32 & 56.00 & 37.29 & 77.24 & 75.47 & 64.11 \\
    & Cosmos3-Super & 71.07 & 54.77 & 56.62 & 8.96 & 74.51 & 72.00 & 56.32 \\
\bottomrule
\end{tabular}

\end{table*}

\begin{figure*}[!h]
\centering
\includegraphics[width=\textwidth]{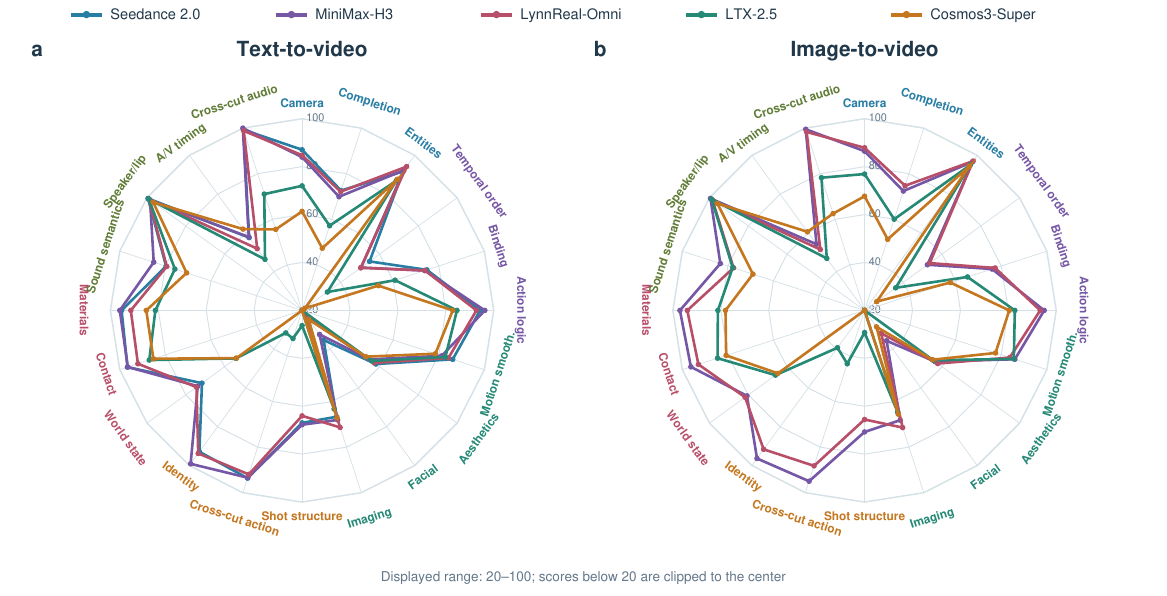}
\caption{Per-metric MSAVP profiles for T2V and I2V. All 20 reported metrics are displayed. The radial scale begins at 20 to reveal differences in the dense upper range; values below 20 are clipped to the center. Seedance 2.0 appears only in the T2V panel.}
\label{fig:msavp-radar}
\end{figure*}

\begin{table*}[!h]
\centering
\caption{Per-metric MSAVP results. Action Binding averages its component and complete-binding subscores. Shot Structure, Cross-shot Identity, and Cross-cut Action use only the 46 multi-shot prompts; Material Behavior pools applicable conditions within each video before averaging the resulting per-video scores. }
\label{tab:msavp-detailed}
\scriptsize
\setlength{\tabcolsep}{3.8pt}
\renewcommand{\arraystretch}{0.92}
\resizebox{\textwidth}{!}{%
\begin{tabular}{@{}lrrrrrrrrr@{}}
\toprule
 & \multicolumn{5}{c}{T2V} & \multicolumn{4}{c}{I2V} \\
\cmidrule(lr){2-6}\cmidrule(l){7-10}
Metric & Seed. & H3 & Lynn & LTX & Cosmos & H3 & Lynn & LTX & Cosmos \\
\midrule
\multicolumn{10}{@{}l}{\textbf{Prompt Fidelity}} \\
Camera Control & \textbf{87.06} & 83.99 & 84.78 & 72.00 & 61.35 & 86.46 & \textbf{87.89} & 76.92 & 67.66 \\
Event Completion & \textbf{72.60} & 69.90 & 72.11 & 57.17 & 47.28 & 72.39 & \textbf{74.71} & 60.03 & 51.19 \\
Entity Fidelity & 91.70 & 92.30 & \textbf{94.19} & 86.90 & 87.63 & 96.62 & \textbf{97.06} & 94.99 & 94.37 \\
\addlinespace[2pt]
\multicolumn{10}{@{}l}{\textbf{Event Execution}} \\
Temporal Order & \textbf{54.80} & 50.45 & 50.25 & 33.06 & 21.50 & 52.43 & \textbf{53.60} & 36.05 & 26.28 \\
Action Binding & \textbf{74.77} & 73.82 & 73.94 & 60.81 & 53.51 & 75.93 & \textbf{77.30} & 65.19 & 57.64 \\
Action Logic & 94.30 & \textbf{96.28} & 92.78 & 84.65 & 82.90 & \textbf{94.97} & 93.30 & 82.71 & 80.38 \\
\addlinespace[2pt]
\multicolumn{10}{@{}l}{\textbf{Visual Quality}} \\
Motion Smoothness & \textbf{86.10} & 81.30 & 84.23 & 82.90 & 78.49 & \textbf{85.96} & 83.71 & 85.50 & 77.46 \\
Aesthetic Quality & \textbf{58.07} & 54.77 & 57.28 & 55.87 & 52.83 & 56.32 & \textbf{57.72} & 55.36 & 55.01 \\
Facial Consistency & \textbf{35.27} & 32.40 & 23.18 & 16.06 & 24.68 & \textbf{35.64} & 31.78 & 18.53 & 28.49 \\
Imaging Quality & 66.67 & 68.07 & \textbf{71.35} & 63.25 & 67.16 & 68.18 & \textbf{71.42} & 64.60 & 65.53 \\
\addlinespace[2pt]
\multicolumn{10}{@{}l}{\textbf{Multi-shot Coherence}} \\
Shot Structure & 66.94 & \textbf{67.71} & 64.00 & 26.33 & 14.03 & \textbf{70.72} & 65.53 & 29.24 & 12.15 \\
Cross-cut Action & \textbf{93.62} & 93.41 & 91.97 & 32.32 & 1.09 & \textbf{94.96} & 88.22 & 43.39 & 8.89 \\
Cross-shot Identity & 92.93 & \textbf{99.15} & 93.77 & 31.60 & 2.17 & \textbf{96.39} & 91.67 & 39.25 & 5.82 \\
\addlinespace[2pt]
\multicolumn{10}{@{}l}{\textbf{World Plausibility}} \\
World-state Consistency & 71.65 & 73.82 & \textbf{74.47} & 54.07 & 53.86 & 80.68 & \textbf{81.65} & 65.97 & 64.70 \\
Contact Response & \textbf{96.64} & 96.47 & 92.04 & 87.07 & 85.27 & \textbf{96.23} & 92.92 & 84.53 & 80.76 \\
Material Behavior & 95.33 & \textbf{96.09} & 91.49 & 81.24 & 85.03 & \textbf{97.07} & 94.00 & 81.23 & 78.06 \\
\addlinespace[2pt]
\multicolumn{10}{@{}l}{\textbf{Audio Alignment}} \\
Sound Semantics & 79.69 & \textbf{85.11} & 79.38 & 75.82 & 70.67 & \textbf{83.30} & 77.50 & 77.87 & 68.99 \\
Speaker--lip Sync & \textbf{99.47} & 98.44 & 99.00 & 99.08 & 96.83 & \textbf{99.50} & 98.50 & 98.92 & 95.94 \\
Action--sound Timing & 57.56 & 57.86 & 51.85 & 46.34 & \textbf{61.96} & 54.14 & 51.43 & 46.83 & \textbf{60.56} \\
Cross-cut Audio Continuity & 99.30 & \textbf{100.00} & 98.83 & 71.05 & 55.56 & \textbf{99.50} & 98.33 & 78.25 & 62.50 \\
\bottomrule
\end{tabular}%
}

\end{table*}

\begin{table*}[t]
\centering
\caption{The 20 reported MSAVP metrics, grouped into six equally weighted capability families. Condition-based metrics use a 0--100\% completion score, while specialist metrics retain their native outputs on a 0--100 scale.}
\label{tab:msavp-metrics}
\scriptsize
\setlength{\tabcolsep}{4pt}
\renewcommand{\arraystretch}{0.92}
\begin{tabular}{@{}p{0.31\textwidth}p{0.63\textwidth}@{}}
\toprule
Metric & What is checked \\
\midrule
\multicolumn{2}{@{}l}{\textbf{Prompt Fidelity}} \\
Camera Control & Requested viewpoint, camera motion, framing, focus, and time effects. \\
Event Completion & Whether each requested event reaches its visible outcome instead of stopping at preparation. \\
Entity Fidelity & Requested objects, counts, visible attributes, text, symbols, and applicable first-frame constraints. \\
\addlinespace[2pt]
\multicolumn{2}{@{}l}{\textbf{Event Execution}} \\
Temporal Order & Explicit before, after, simultaneous, and during relations between events. \\
Action Binding & Correct components and complete actor--action--target tuples, including a specified body part. \\
Action Logic & Whether observed actions start, respond, progress, and end coherently. \\
\addlinespace[2pt]
\multicolumn{2}{@{}l}{\textbf{Visual Quality}} \\
Motion Smoothness & Frame-weighted VMBench MSS over the detected shots~\citep{ling2025vmbench}. \\
Aesthetic Quality & WBench Aesthetic Quality: mean CLIP--LAION aesthetic prediction from frames sampled at 2 fps~\citep{ying2026wbench}. \\
Facial Consistency & Stability and expected count of detectable primary faces under Facial-100~\citep{zhou2026avgenbench}. \\
Imaging Quality & WBench Imaging Quality: mean MUSIQ perceptual-quality prediction from frames sampled at 2 fps~\citep{ying2026wbench}. \\
\addlinespace[2pt]
\multicolumn{2}{@{}l}{\textbf{Multi-shot Coherence}} \\
Shot Structure & Requested shot count, content allocation, order, and event-triggered cuts. \\
Cross-cut Action & Action phase, contact, held objects, and direction across each applicable observed cut. \\
Cross-shot Identity & Identity of people, animals, and key objects across shots. \\
\addlinespace[2pt]
\multicolumn{2}{@{}l}{\textbf{World Plausibility}} \\
World-state Consistency & Object permanence, lasting interaction results, layout, and connected space. \\
Contact Response & Contact timing, force response, collision direction, penetration, and action at a distance. \\
Material Behavior & Behavior of rigid, brittle, liquid, rope, chain, cloth, and other visible materials. \\
\addlinespace[2pt]
\multicolumn{2}{@{}l}{\textbf{Audio Alignment}} \\
Sound Semantics & Requested dialogue, interaction sounds, ambience, music, and silence. \\
Speaker--lip Sync & Correct speaker, absence of false lip motion, and visible speech alignment. \\
Action--sound Timing & Whether a sound belongs to the same visible event and occurs at the expected time. \\
Cross-cut Audio Continuity & Dialogue, voice, and acoustic continuity or justified change at each observed cut. \\
\bottomrule
\end{tabular}
\end{table*}

\paragraph{Step 1: freeze the scoring checklist.}
The first step converts each prompt into explicit, reusable scoring items. We use GPT-6 Astra through Codex with medium reasoning for the model-based evaluation stages because its visual reasoning supports more general and reliable semantic judgments than narrowly designed specialist models. To construct the base checklist, Astra reads the T2V prompt without seeing any generated video and returns a structured set of conditions. For I2V, it adds only conditions that are directly visible in the shared opening image. This image-derived overlay may introduce opening-state constraints for entities and attributes, cross-shot identity, and spatial or world state, but it cannot redefine action bindings, events, temporal relations, materials, or audio requirements.

Freezing the checklist ensures that every model receives the same test. All candidate models evaluated under the same prompt and generation mode use this exact set. After viewing a candidate video, the evaluator may judge whether a condition is satisfied, but it cannot merge, delete, rewrite, or invent conditions.

\paragraph{Step 2: collect specialist evidence.}
The second step extracts measurements that are more reliable when produced by dedicated tools. TransNetV2~\citep{soucek2020transnetv2} proposes shot boundaries at a threshold of 0.1, after which Codex examines the video and rejects flashes, occlusions, and rapid camera motion that resemble cuts.

The audio specialist reports what is actually audible and when each sound occurs. Qwen3-Omni-30B-A3B-Thinking~\citep{xu2025qwen3omni} reads the original soundtrack with 4-fps visual sampling. It identifies dialogue, interaction sounds, ambience, and music without treating a visible event as evidence that the corresponding sound is present. Following WBench~\citep{ying2026wbench}, we report Aesthetic Quality using CLIP ViT-L/14~\citep{radford2021clip} with the LAION linear aesthetic head~\citep{laion2022aesthetic}, and Imaging Quality using MUSIQ~\citep{ke2021musiq}; both are averaged over frames sampled at 2 fps. We additionally apply MSS, the VMBench Motion Smoothness Score~\citep{ling2025vmbench}, to each detected shot and aggregate the resulting scores by evaluated frames. Facial-100 is our percentage-scale name for the AVGen-Bench Facial Consistency metric~\citep{zhou2026avgenbench}; following that benchmark, it uses InsightFace features with face tracking and identity clustering to measure the stability and expected count of detectable primary faces. Videos without an applicable detected face are excluded from this metric rather than treated as failures.

\paragraph{Step 3: run the prompt-blind judge.}
The third step isolates judgments that should depend only on the generated video. A separate Codex context receives the candidate video and proposed cut locations, but not the prompt, opening image, model name, audio analysis, or other scores. It evaluates action continuity across the observed cuts. Consequently, a requested action that never appears remains a semantic failure rather than being counted again as a physical violation.

\paragraph{Step 4: run the main judge.}
The fourth step scores the 15 checklist-based metrics against the frozen checklist and specialist evidence. All condition-based metrics are expressed on a 0--100 scale. This judge context receives the candidate video, original prompt, applicable checklist, and specialist outputs, but not the prompt-blind judgment. The judge can inspect the native video, revisit short intervals, and examine adjacent frames when small objects, physical contact, material response, or cut boundaries require closer observation. Its output contains one decision and a short supporting evidence interval for every applicable condition.

\paragraph{Step 5: validate and aggregate.}
The final step makes the model output mechanically checkable. After validating the item-level decisions, the pipeline computes all aggregate scores directly rather than accepting totals supplied by the judge.

\paragraph{Scoring and reporting.}
Condition-based metrics use three observable outcomes. For condition $k$ of metric $m$ on video $i$, $c_{imk}$ is 1 for satisfied, $\frac{1}{2}$ for substantively partially satisfied, and 0 for failed. The metric score is
\begin{equation}
 s_{im}=\frac{100}{|\mathcal C_{im}|}\sum_{k\in\mathcal C_{im}}c_{imk}.
\end{equation}
Partial credit requires visible partial success and cannot be used merely to represent uncertainty. A requested event that never appears remains a failed semantic condition. Shot structure, cross-shot identity, and cross-cut action continuity are evaluated only on the 46 prompts that explicitly request multiple shots. If such a prompt produces a single-shot video, cross-shot identity and cross-cut action continuity receive zero, while shot structure is scored against its requested shot conditions. If the generated video has multiple shots but no action spans a cut, action continuity receives 100 because no cross-cut action discontinuity is present. The 54 single-shot prompts do not enter the model-level denominators of these three multi-shot metrics. A predeclared case without dialogue receives 100 for speaker--lip synchronization and remains distinct from an evaluator failure. For cross-cut audio continuity, a single-shot caption receives 100, whereas a caption that requests multiple shots receives 0 if the generated video remains single-shot. Component compliance and complete actor--action--target binding compliance are averaged into one binding metric. For material behavior, all applicable material conditions within each video are pooled into a per-video score, and the model-level metric is the mean of these per-video scores.

We first average metrics within six capability families: Prompt Fidelity (three metrics), Event Execution (three), Visual Quality (four), Multi-shot Coherence (three), World Plausibility (three), and Audio Alignment (four), exactly as grouped in Table~\ref{tab:msavp-metrics}. Prompt Fidelity measures requested entities, completed events, and camera control. Event Execution combines actor--action--target binding, temporal relations, and the internal logic of observed action chains. Visual Quality combines frame-level aesthetics and imaging quality with motion smoothness and facial consistency. Multi-shot Coherence is restricted to explicitly multi-shot prompts and measures requested shot structure, entity identity, and action continuity across cuts. World Plausibility measures persistent world state, contact response, and material behavior. If $\mathcal M_f$ denotes the metrics in family $f$, its score is $g_f=|\mathcal M_f|^{-1}\sum_{m\in\mathcal M_f}\bar{s}_m$. The overall score gives every family equal weight,
\begin{equation}
 \mathrm{MSAVP}=\frac{1}{6}\sum_{f\in\mathcal F}g_f,\qquad |\mathcal F|=6.
\end{equation}
This hierarchy prevents a family from receiving more weight merely because it contains more metrics. 

We report MSAVP scores for our real-time world model LynnReal-Omni, Seedance 2.0~\citep{seed2026seedance20}, MiniMax-H3~\citep{minimax2026h3}, LTX-2.5~\citep{lightricks2026ltx25}, and Cosmos3-Super~\citep{nvidia2026cosmos3}, keeping T2V and I2V results separate. Aesthetic quality and imaging quality retain their native specialist scales. GPT-6 Astra through Codex constructs the frozen checklists and supplies the model-based judgments. 

\subsubsection{Benchmark results}

Table~\ref{tab:msavp-overview} reports the six family scores and their equal-weight average. In T2V, Seedance 2.0 obtains the highest overall score at 79.39, narrowly ahead of MiniMax-H3 at 79.27. LynnReal-Omni reaches 77.76 in T2V and 79.20 in I2V, 1.63 points behind the best T2V system and 1.45 points behind MiniMax-H3 in I2V. Unlike the offline generators in the comparison, LynnReal-Omni is designed as a real-time world model. It leads I2V Prompt Fidelity and is particularly strong in entity fidelity, world-state consistency, aesthetic quality, and imaging quality. These results show that its low-latency interactive design retains competitive instruction and visual-world fidelity rather than trading them away wholesale for speed. Seedance leads Prompt Fidelity, Event Execution, and Visual Quality in T2V while MiniMax-H3 remains strongest in overall I2V, Multi-shot Coherence, World Plausibility, and Audio Alignment. LTX-2.5 and Cosmos3-Super underperform across most instruction-intensive and multi-shot measures. The per-metric profiles are visualized in Figure~\ref{fig:msavp-radar}; Table~\ref{tab:msavp-detailed} gives every reported metric.

\subsection{Lightweight VAE Evaluation}
  \label{sec:vae-evaluation}

  We evaluate native-768P reconstruction quality using the official
  decoder and the selected EMA lightweight decoder with native or
  adaptive tiling. Visual comparisons use matched source frames and
  crop coordinates, while quantitative metrics are computed from
  uncompressed RGB outputs before file encoding.

  \begin{figure}[!h]
  \centering
  \includegraphics[width=\linewidth]{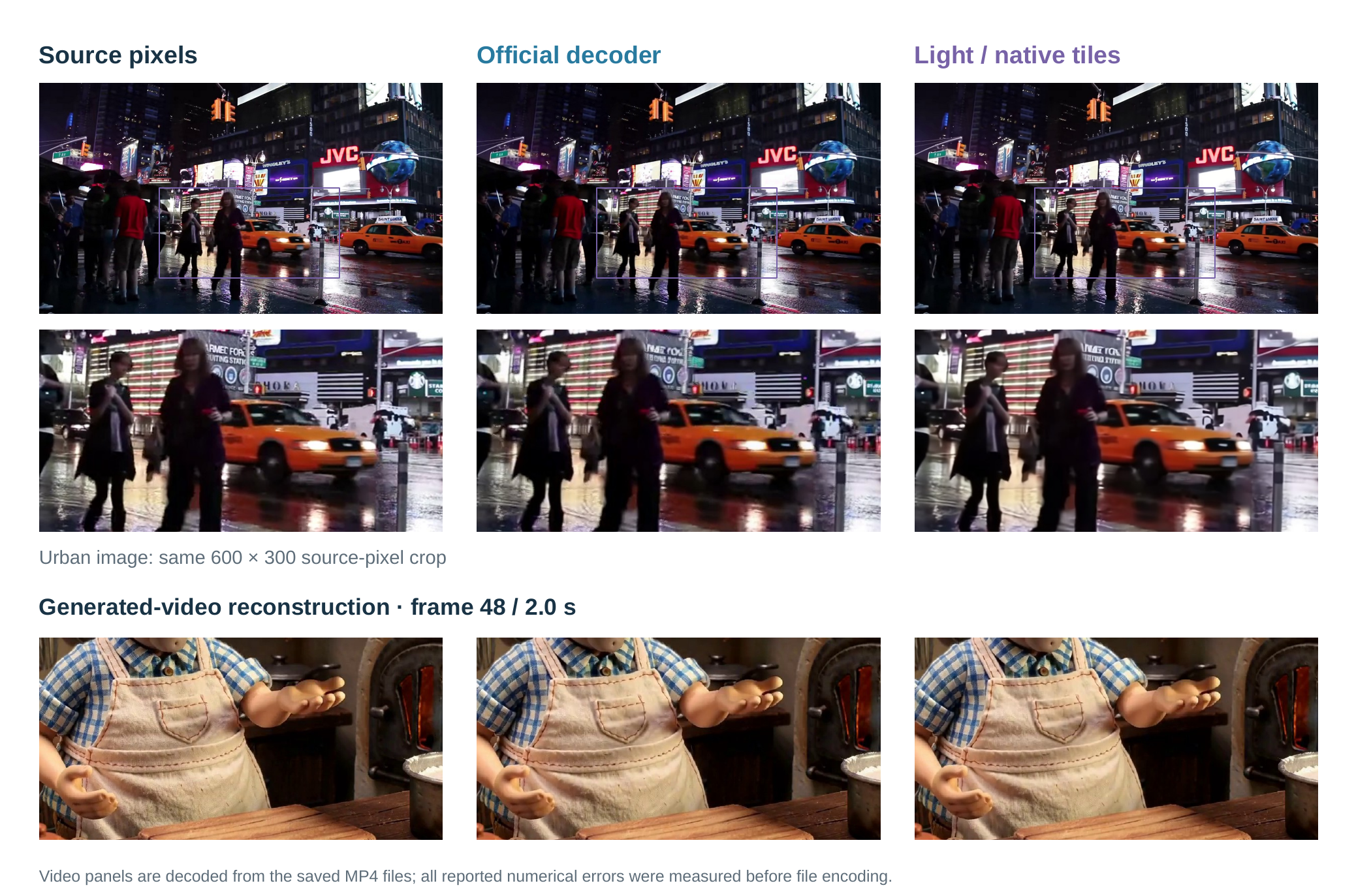}
  \caption{
  Native-768P reconstruction comparisons. The urban image and its
  enlarged crop use the same five-token image context across methods.
  The clay-video crop uses identical source frames and spatial
  coordinates. No visual enhancement is applied. Video panels are
  extracted from saved MP4 files; quantitative metrics are computed
  before file encoding.
  }
  \label{fig:vae-image}
  \end{figure}

  \paragraph{Spatial fidelity.}
  Figure~\ref{fig:vae-image} shows that the lightweight decoder preserves
  the overall scene layout, clothing patterns, and table textures,
  although fine details can be softened. Its higher reconstruction
  PSNR on the urban image indicates lower pixel error for that example,
  but does not establish better perceptual quality on generated videos.
  The enlarged crops complement aggregate metrics by revealing local
  differences in texture and edge sharpness.

  \begin{figure}[!h]
  \centering
  \includegraphics[width=\linewidth]{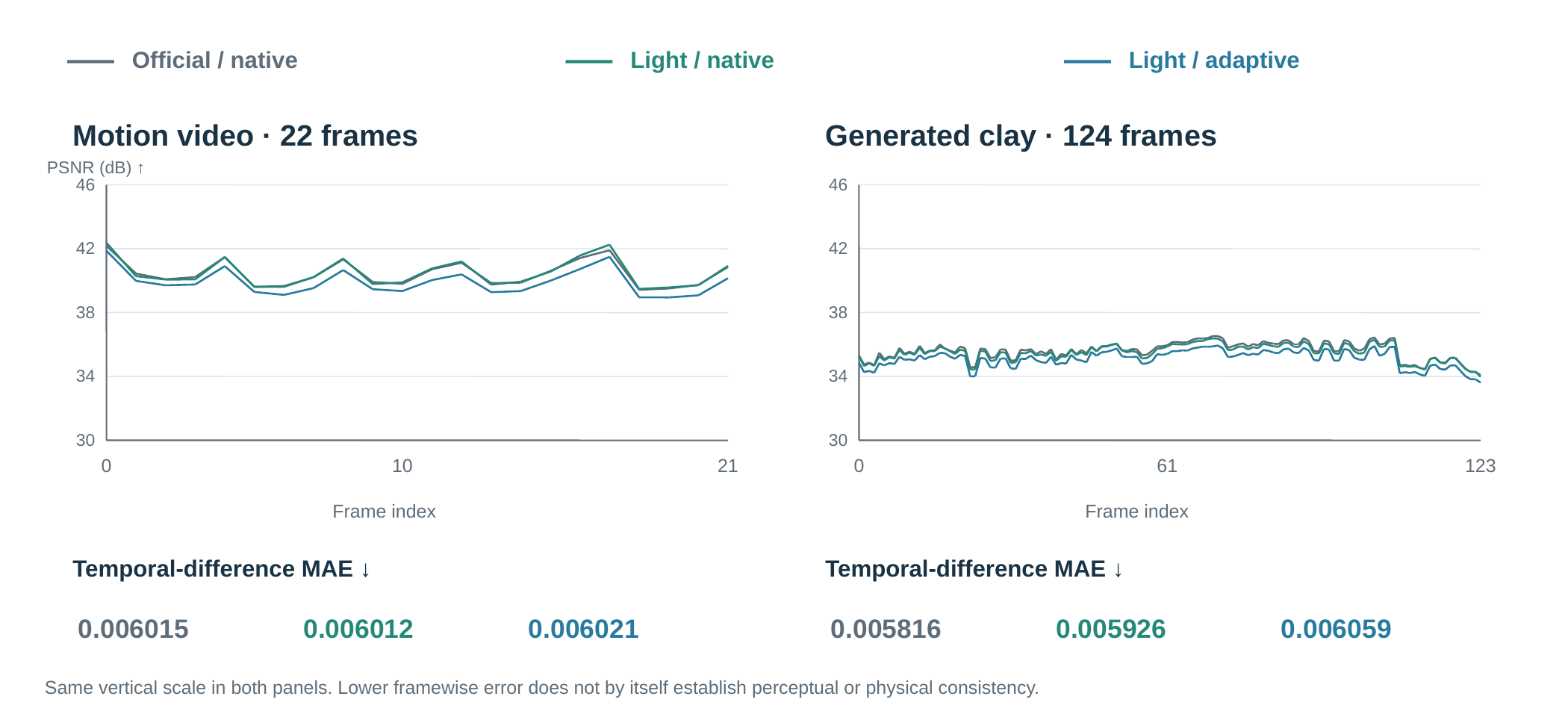}
  \caption{
  Per-frame PSNR and temporal-difference MAE on two native-768P video
  inputs. All decoded frames are included, with consistent vertical
  ranges across clips for each metric. Curves and aggregate values
  are computed from uncompressed RGB reconstructions against the
  same source videos.
  }
  \label{fig:vae-temporal768}
  \end{figure}

\paragraph{Temporal fidelity.}
Figure~\ref{fig:vae-temporal768} exposes framewise variations that clip-averaged scores can obscure. On the motion clip, temporal-difference MAE is nearly identical for the official and native-tile lightweight decoders: 0.006015 and 0.006012, respectively. On the clay clip, the official decoder achieves lower error, 0.005816 versus 0.005926. Adaptive lightweight decoding yields 0.006021 and 0.006059 on the two clips, respectively, slightly above the corresponding native-tile results.

The framewise curves follow broadly similar trends, including a late quality drop on the clay clip shared by all three configurations. This behavior suggests that the observed fluctuations cannot be attributed solely to reduced decoder depth. Overall, native-tile lightweight decoding maintains temporal reconstruction errors close to the official decoder on these inputs, with remaining differences in fine-detail preservation. These results characterize reconstruction fidelity;

\subsection{Inference Latency}
\label{sec:inference-latency}
We measure warm inference on single H100 80GB GPUs, using two warmup
calls and five measured calls per configuration, repeated on a second GPU.
Prompt and seed are fixed within each setting. Standard and Flash use
four and three denoiser evaluations, respectively. CUDA events measure
the DiT and video decoder; synchronized generation wall time additionally
includes scheduling, audio decoding, RGB conversion, and output transfer.
Conditioning, loading, file encoding, and warmup compilation or tuning
are excluded. Tables report pooled medians of ten calls.

\paragraph{Component ablations.}
Table~\ref{tab:latency-components} isolates successive execution changes
at 540p and 22 frames. Each model's complete sequence of configurations
runs in one process on the same GPU. The standard baseline uses BF16;
Flash retains its trained INT8 checkpoint throughout. QKV concatenation
is measured separately before standard-model quantization. Subsequent
rows add one component to the preceding configuration. The lightweight
decoder row includes its batched implementation; tile geometry and
compilation are then varied separately.

\begin{table*}[!ht]
\centering\small
\setlength{\tabcolsep}{5pt}
\begin{tabular}{lrrrrrr}
\toprule
& \multicolumn{3}{c}{Standard (4 NFE)} & \multicolumn{3}{c}{Flash (3 NFE)} \\
\cmidrule(lr){2-4}\cmidrule(lr){5-7}
Component added & DiT+dec. & Generate & GiB & DiT+dec. & Generate & GiB \\
\midrule
Baseline & 2345 & 2682 & 77.2 & 1592 & 1920 & 52.4 \\
+ QKV concatenation & 2355 & 2684 & 77.2 & --- & --- & --- \\
+ W8A8 projections & 3292 & 3627 & 59.9 & --- & --- & --- \\
+ operator fusion & 1719 & 2045 & 59.9 & 976 & 1309 & 52.4 \\
+ autotuned Triton INT8 GEMM & 1432 & 1767 & 59.9 & 869 & 1206 & 52.4 \\
+ FlashAttention 3 & 1300 & 1633 & 59.9 & 823 & 1155 & 52.4 \\
+ time-modulation cache & 1266 & 1600 & 60.0 & 801 & 1140 & 52.5 \\
+ GPU RGB conversion & 1268 & 1370 & 60.0 & 801 & 903 & 52.5 \\
+ light decoder (native tiles) & 1081 & 1175 & 54.9 & 599 & 679 & 47.4 \\
+ adaptive decoder tiles & 935 & 1029 & 53.9 & 460 & 557 & 46.4 \\
+ decoder compilation & 843 & 959 & 53.9 & 377 & 479 & 46.4 \\
\bottomrule
\end{tabular}
\caption{Cumulative component ablation on H100, 22 frames at 540p.
Latencies are in milliseconds; memory is peak allocated GiB. Flash
already uses INT8 and concatenated QKV in its baseline. FA3 changes
attention in both DiT and VAE. DiT+dec. is the median of per-call sums.
Component effects are compared within each model, without attributing
the difference between separately trained models to a single design choice.}
\label{tab:latency-components}
\end{table*}

Unfused INT8 reduces memory but increases standard-model latency. Operator fusion reduces generation wall time by 43.6\% for standard and 31.6\% for Flash, computed from within-GPU comparisons. Triton GEMM, time-modulation caching, and GPU RGB conversion retain
exact paired video latents and RGB in these tests. Quantization, operator fusion,
attention changes, and decoder replacement change numerical outputs;
latency gains alone do not establish equivalent visual quality.
Decoder compilation changes RGB by at most one 8-bit intensity level.

Two further controls separate architectural and scheduling effects. With the same Flash weights and three evaluations, disabling token compression increases DiT time from 263 to 453\,ms; this control is evaluated without retraining. With identical latents, native tile geometry, batched scheduling, precision, and eager execution, replacing the 36-block decoder with the 26-block student reduces decoding from 469 to 340\,ms.

\paragraph{Resolution and duration.}
Table~\ref{tab:latency-verified} reports the complete accelerated path:
W8A8, FA3, Triton GEMM, time-modulation caching, GPU RGB conversion,
and the compiled lightweight decoder with adaptive tiles. The 540p
rows reuse the final component configurations; 768p results are separate
measured runs. All outputs use 24\,fps. The 540p canvas is cropped from
$960\times544$ to $960\times540$; 768p uses $1344\times768$.
Native temporal padding is included in computation. The 15\,s test
explicitly permits 362 computed frames and retains the first 360;
64-bit kernel indices support tensors exceeding $2^{31}$ elements.
Each video is generated as one clip.

\begin{table*}[!ht]
\centering\small
\setlength{\tabcolsep}{5pt}
\begin{tabular}{llcrrrrr}
\toprule
Output & Model & NFE & DiT & Decoder & DiT+dec. & Generate & Peak \\
\midrule
540p, 22 frames & Standard & 4 & 0.724 & 0.119 & 0.843 & 0.959 & 53.9 \\
540p, 22 frames & Flash & 3 & 0.262 & 0.115 & 0.377 & 0.479 & 46.4 \\
\midrule
768p, 5 s & Standard & 4 & 18.293 & 1.723 & 20.018 & 20.514 & 58.0 \\
768p, 5 s & Flash & 3 & 5.474 & 1.724 & 7.198 & 7.661 & 49.9 \\
\midrule
768p, 10 s & Standard & 4 & 56.969 & 3.428 & 60.393 & 61.490 & 63.0 \\
768p, 10 s & Flash & 3 & 16.322 & 3.394 & 19.716 & 20.763 & 54.4 \\
\midrule
768p, 15 s & Standard & 4 & 116.207 & 5.112 & 121.315 & 122.920 & 68.1 \\
768p, 15 s & Flash & 3 & 33.505 & 5.163 & 38.665 & 40.362 & 58.9 \\
\bottomrule
\end{tabular}
\caption{Measured resolution and duration scaling. Latencies are in
seconds; peak allocated memory is in GiB. Each row aggregates five
measured calls on each of two GPUs. DiT+dec. is the median of per-call
sums, which can differ from the sum of component medians.}
\label{tab:latency-verified}
\end{table*}

These measurements characterize warm execution rather than cold-start
or request-to-display latency. Original videos and numerical comparisons
are retained for inspection; long-clip latencies are measured directly.

\paragraph{Packaged inference validation.}
A separate single-H100 check uses the public INT8 configuration with the
compiled lightweight decoder and adaptive tiles. After two preparation calls,
we report medians of five measured calls (Table~\ref{tab:latency-install}).
The installer prepares common shapes and persists kernel caches; first-use
compilation is recorded separately from measured inference. Grouped INT8 GEMM
and residual--normalization fusion preserve paired latents, audio, and RGB
within each tested attention configuration. Changing attention or decoder tile
geometry is a separate numerical change. These checks do not establish
bitwise equivalence across backends.

\begin{table}[!ht]
\centering\small
\begin{tabular}{llrr}
\toprule
Model & Attention & DiT+dec. & Generate \\
\midrule
Standard & FA3 & 857 & 957 \\
Flash & FA3 & 383 & 477 \\
Standard & FA2 & 990 & 1088 \\
Flash & FA2 & 422 & 518 \\
Standard & cuDNN & 950 & 1057 \\
Flash & cuDNN & 416 & 516 \\
\bottomrule
\end{tabular}
\caption{Installed inference on one H100, 540p and 22 frames. Times are
in milliseconds, with four Standard or three Flash evaluations. The
light decoder and tile layout are fixed. Each backend is measured in a
separate process; these are deployment checks rather than isolated
attention-kernel speedups.}
\label{tab:latency-install}
\end{table}

The FA3 path takes approximately 0.86\,s for Standard and 0.38\,s for Flash
for denoising plus video decoding. Independent public-command runs give
855 and 381\,ms, respectively (940 and 458\,ms generation wall time),
with five measured calls after two warmups. Public FA2 runs give
986 and 420\,ms for DiT plus decoder. Without FA3, the measured Standard path
remains above 0.8\,s; a successful fallback is not evidence of equal latency.
Generation wall time is reported separately and includes additional pipeline
work. Installation, conditioning, warmup, and file encoding are excluded.

\begin{figure}[!h]
\centering
\includegraphics[width=0.8\textwidth]{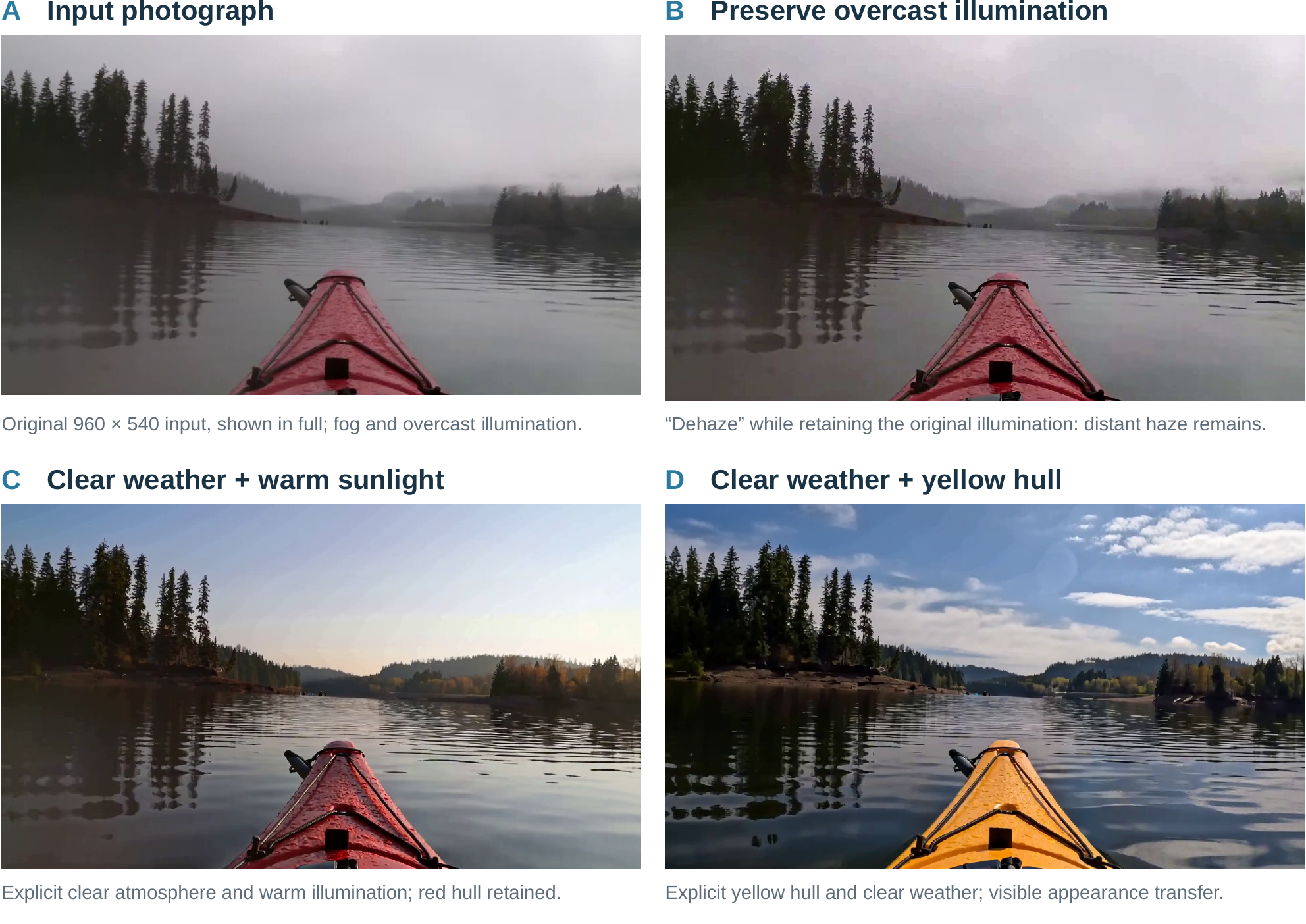}
\caption{Fixed-model prompt ablation.}
\label{fig:weather}
\end{figure}
\subsection{Reference-guided weather and appearance editing}
Figure~\ref{fig:weather} isolates the effect of specifying an edited visual state. The original lake photograph is a 960$\times$540 picture reference; every generated output is natively 1344$\times$768. We fix the reference-capable base, active EMA adapter, seed, attention backend, four denoiser evaluations, and 22-frame request. Each prompt uses the same ordered picture-reference layout and asks for the completed edit from the opening frame with a locked camera. The exported image is frame zero, and the complete generated clip is retained for temporal inspection.

A conservative instruction to remove haze while preserving overcast illumination leaves substantial distant haze. Describing transparent air, distinct wooded hills, warm sunlight, and a pale blue sky produces a visibly clearer scene while retaining the red kayak. A separate prompt explicitly changes the hull to yellow and also requests clear weather. The black cords and main viewpoint remain recognizable in this example. Sampled frames across the short companion clips preserve the principal composition and edited hull color, although water ripples continue to change. These are controlled prompt examples, not a benchmark-wide estimate of editing reliability.

The broader edits change illumination and synthesize detail hidden by the source fog. They therefore demonstrate generative weather and appearance editing, rather than recovery of a paired clear target. We do not report restoration PSNR or infer improved video physics from a successful color change. A separate native768p nighttime video test retains visible rain near the lamp under the conservative removal instruction.

\begin{figure}[p]
\centering
\includegraphics[width=0.9\textwidth]{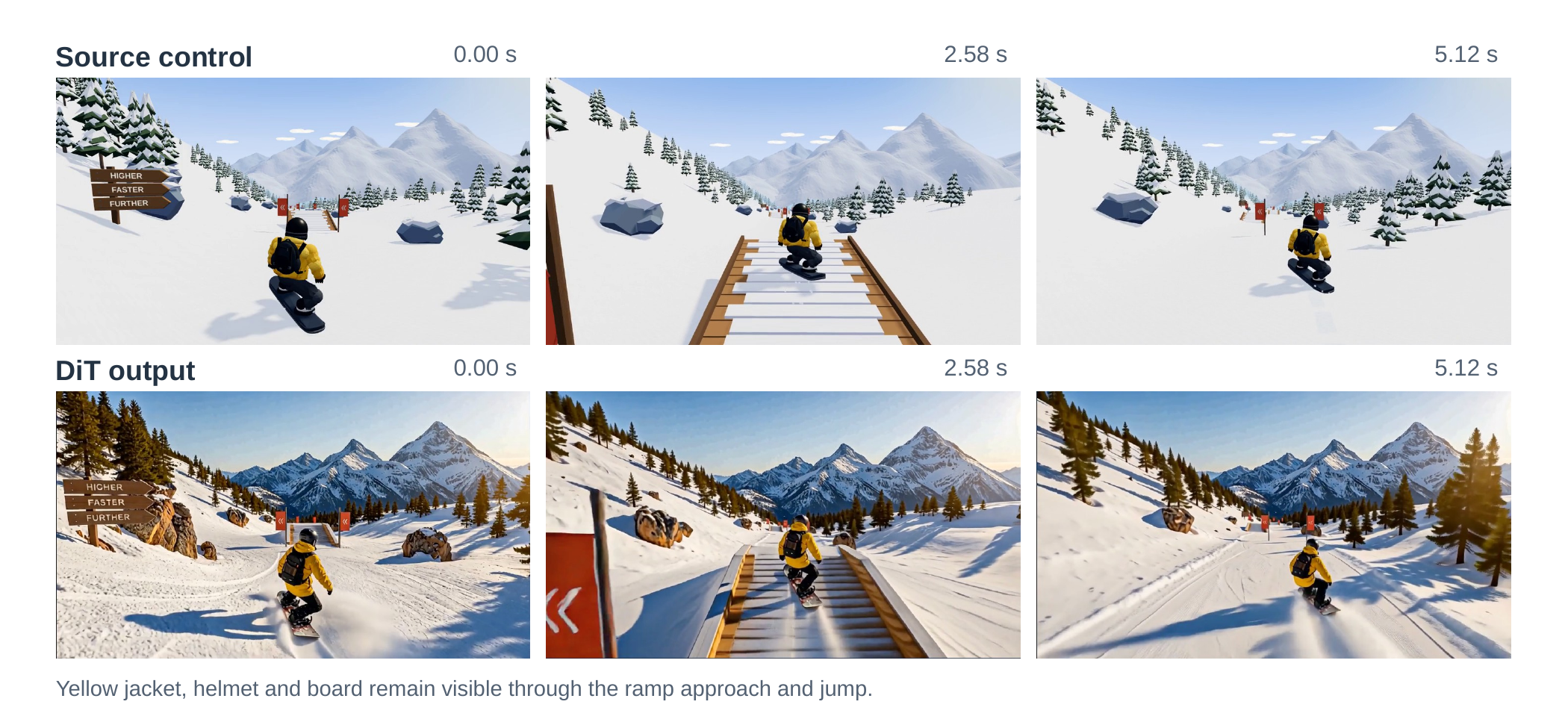}
\caption{Recorded game control and anime-conditioned generation at matched timestamps. The source recording supplies the motion timeline; a locally edited first frame supplies appearance. This development example uses four denoiser evaluations and does not measure live game input-to-display latency.}
\label{fig:game-comparison}
\end{figure}

\begin{figure}[p]
\centering
\includegraphics[width=\textwidth]{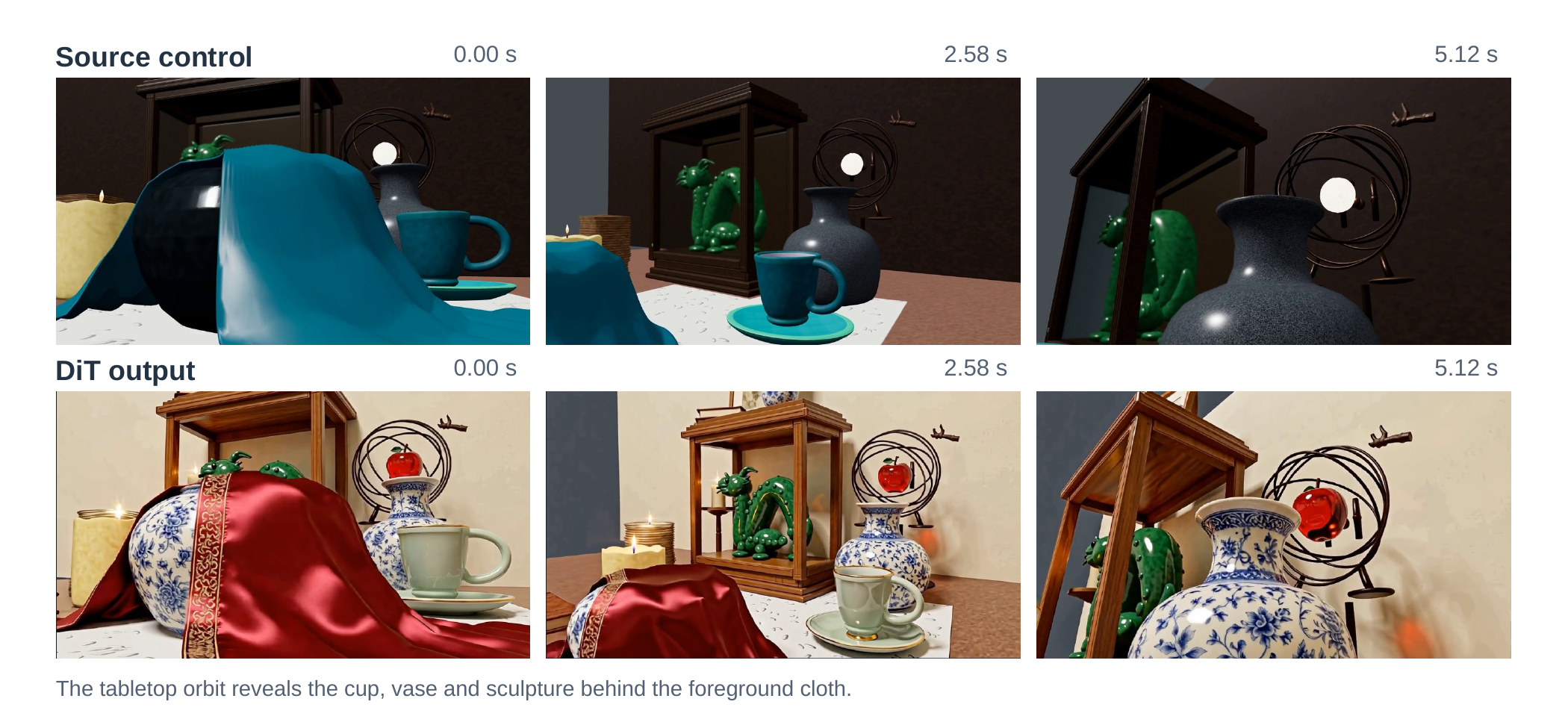}
\caption{An existing mesh camera tour and its clay-conditioned video output at matched timestamps. Foreground occlusion and parallax expose objects behind the cloth. Changes to surface appearance do not imply exact geometric reconstruction; the displayed frames are qualitative development evidence.}
\label{fig:mesh-comparison}
\end{figure}

\section{Conclusion}
\method{} couples native multi-modal video diffusion with explicit controls over appearance, geometry, motion, and history. Its data pipeline separates physical timing from semantic grouping while verifying every conditioning asset against its source; a shared standard denoiser supports multiple input layouts, and bounded-history long-video generation maintains a fixed memory interface. Agent-produced scenes and executable games further provide inspectable controls whose visual realization can be altered through reference-guided generation. The experiments disentangle several questions that might otherwise be conflated—exact reload fidelity, quality under compression, decoder context, warm throughput, long-rollout continuity, and responsiveness to edits—showing that current warm 540p measurements demonstrate useful acceleration but do not establish equivalent native 768p speed or live-interaction latency. Likewise, a plausible image edit does not prove repaired video physics, and catalog or benchmark design counts are no substitute for measured training exposure or completed evaluations; native 768p development comparisons and multi-modal demonstrations therefore require their own source-linked generation and temporal review. These distinctions make the report's claims testable as the release and evaluations develop. Future evaluation should complete independent cross-model MSAVP scoring, measure human agreement with automated judges, and test longer interactive trajectories with recorded action-to-display timing, while improving temporal stability under strong style transfer and preserving physical contacts remain central challenges for using a fast video generator as a controllable visual renderer.

\section{Contributors}
The authors are listed in descending order of their actual contributions as follows:

\begin{center}
Xiaofeng Mao\textsuperscript{1,2,3,*,$\ddagger$}\\
Peijia Lin\textsuperscript{1,2,*}\\
Shaohao Rui\textsuperscript{1,2,3,*,$\ddagger$}\\
Yibo Zhang\textsuperscript{1,2,*}\\
Haibin Wan\textsuperscript{1,2}\\
Weijie Ma\textsuperscript{1,2,4,$\dagger$}
\end{center}

\begin{center}
\textsuperscript{1}LynnReal Lab \quad
\textsuperscript{2}Shanghai Innovation Institute \quad
\textsuperscript{3}Shanghai Jiao Tong University \quad
\textsuperscript{4}Fudan University
\end{center}

\begin{center}
\textsuperscript{*}Equal contribution. \quad
\textsuperscript{$\ddagger$}Project Lead. \quad
\textsuperscript{$\dagger$}Corresponding Author.
\end{center}

\clearpage
\bibliographystyle{plainnat}
\bibliography{references}
\end{document}